%% file: main.tex
\pdfoutput=1
\documentclass{article} %
\usepackage{iclr2027_conference,times}

\input{math_commands.tex}

\usepackage[utf8]{inputenc} %
\usepackage[T1]{fontenc}    %
\usepackage{hyperref}       %
\usepackage{url}            %
\usepackage{booktabs}       %
\usepackage{amsfonts}       %
\usepackage{nicefrac}       %
\usepackage{microtype}      %
\usepackage{xcolor}         %
\usepackage{booktabs}
\usepackage{multirow}
\usepackage{array}
\usepackage{amsmath}
\usepackage{graphicx}
\usepackage{subcaption}
\usepackage{wrapfig}
\usepackage{amsmath}

\title{Simply Stabilizing the Loop via Fully Looped Transformer}

\author{Rao Fu$^{1}$, Zixuan Yang$^{2}$, Jiankun Zhang$^{2}$, Jing Ma$^{1}$\thanks{Correspondence to \texttt{majing@hkbu.edu.hk}}, Hechang Chen$^{2}$, Yu Li$^{2}$, Yi Chang$^{2}$ \\
$^{1}$Hong Kong Baptist University, $^{2}$Jilin University
}

\iclrfinalcopy %
\begin{document}

\maketitle
\lhead{Preprint}

\input{Sections/abstract}
\input{Sections/introduction}

\input{Sections/relatedworks}

\input{Sections/understand}

\input{Sections/architecture}

\input{Sections/experiment}

\input{Sections/conclusion}

\clearpage
\subsection*{AI use statement}

We used generative AI tools for language polishing, figure preparation, and assistance in interpreting experimental results. All AI-assisted content has been reviewed and verified. The authors take full responsibility for the final manuscript.

\subsection*{Reproducibility statement}

We document the training data, optimization settings, and hardware in Appendix~\ref{app:train}, the model architecture and attention variants in Appendix~\ref{app:arc}, and the evaluation protocols, including test-time compute adaptation, in Appendix~\ref{app:exp}. These details complement the experimental setup in Section~\ref{sec:exp:setup} and specify the configurations used to obtain the reported results.

\bibliography{iclr2027_conference}
\bibliographystyle{iclr2027_conference}

\appendix
\input{Sections/appendix}

\end{document}

%% file: math_commands.tex
\usepackage{amsmath,amsfonts,bm}

\def\eqref#1{equation~\ref{#1}}

\def\1{\bm{1}}

\DeclareMathAlphabet{\mathsfit}{\encodingdefault}{\sfdefault}{m}{sl}
\SetMathAlphabet{\mathsfit}{bold}{\encodingdefault}{\sfdefault}{bx}{n}



%% file: Sections/abstract.tex
\begin{abstract}

Scaling model performance typically requires increasing model size. Looped Transformer (LT) offers a parameter-efficient alternative by repeatedly reusing the same Transformer blocks, converting additional computation into performance gains without increasing the parameter count or context length. However, increasing the number of loop iterations makes LT increasingly difficult to optimize, often leading to gradient oscillation, residual explosion, and training collapse. We propose the Fully Looped Transformer (FLT), a simple, parameter-free redesign that addresses these problems without adding learnable parameters or imposing specially designed optimization constraints. FLT combines a Fully Looped Architecture, which exposes the recurrent state to every layer, with Attention Injection, which controls recurrent information flow through the existing attention module. These modifications enable FLT to train stably through 12 loops in our tested configurations, while baseline looped models enter high-loss regimes. In the Base-size, 6-loop setting, FLT improves downstream-task accuracy by 14.9\% relative to the original LT. Fixed-loop checkpoint evaluations further show that performance improves as inference depth approaches the trained depth, even without randomized loop-count training, revealing a depth-aware test-time compute interface. Extending this interface beyond the trained range is a promising research direction.

\end{abstract}

%% file: Sections/introduction.tex
\section{Introduction}

Large language models (LLMs) demonstrate excellent generalization capabilities across numerous downstream tasks through pretraining on massive text corpora from the internet. However, this scaling paradigm is increasingly constrained by the limited supply of high-quality public text data. According to the Chinchilla scaling law~\citep{hoffmann2022training}, model parameters and training data should scale proportionally. Simply increasing model size without a corresponding increase in data leads to suboptimal training results. Meanwhile,~\citet{villalobos2024position} show that the stock of public human-generated text data available for LLM pretraining grows at only 10\% per year,  while the dataset sizes used in practice have been growing at approximately 2.4$\times$ per year. This growing mismatch between data availability and computational capacity calls for new paradigms that can convert excess compute into performance gains without relying on ever-larger datasets.

These data constraints motivate growing interest in improving model performance through additional computation, rather than relying solely on larger training corpora. Test-time scaling explores this direction by allocating more computation at inference time. A prominent approach uses reinforcement learning~\citep{openai2024openaio1card, guo2025deepseek} to elicit extended chain-of-thought reasoning~\citep{wei2023chainofthoughtpromptingelicitsreasoning}, allowing models to devote more computation to a problem by generating longer reasoning traces before producing a final answer.

Rather than extending the token sequence, additional computation can also be allocated to iterative refinement of hidden representations. Looped Transformer~\citep{giannou2023looped} follows this direction by iteratively reusing the same Transformer backbone blocks. Instead of increasing the model parameter count or expanding the context window, it unrolls the same shared blocks for multiple loop iterations. By virtue of the looping mechanism, Looped Transformer is parameter-efficient~\citep{saunshi2025reasoning}, naturally compatible with test-time compute adjustment~\citep{koishekenov2025encode} and excels at reasoning tasks~\citep{saunshi2025reasoning}.

However, despite these advantages, Looped Transformer remains difficult to train when the number of loop iterations becomes large~\citep{geiping2025scaling, zhu2025scaling}. We also observe the same issue in our experiments: simply increasing the loop iterations of Looped Transformer can lead to training collapse. This motivates us to examine the training dynamics of Looped Transformer. Our diagnosis identifies two instability patterns. First, gradients can oscillate strongly during the early stage of training. Second, the residual-state norm can grow rapidly as loop iterations increase. We refer to these two phenomena as gradient oscillation and residual explosion, respectively.

\begin{figure}[t!]
    \centering
    \includegraphics[width=\linewidth]{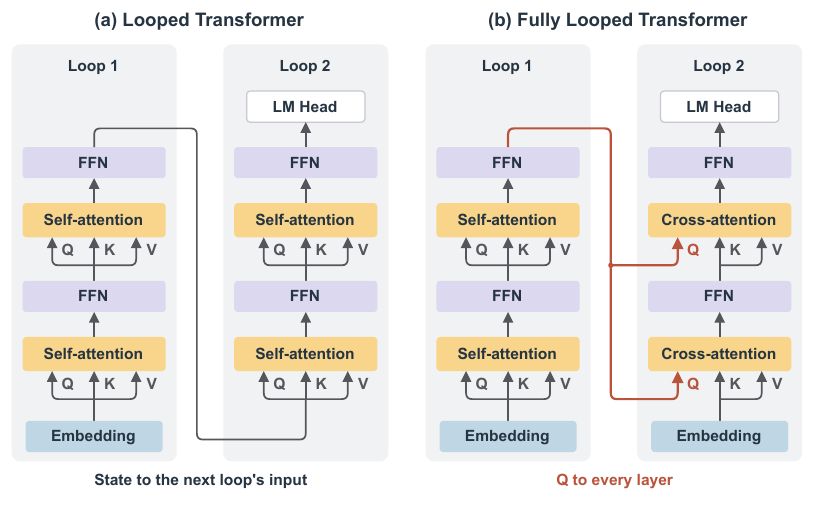}
    \caption{
        \textbf{FLT makes the previous loop state available to every layer through attention.}
        (a) LT passes the previous loop output only to the next loop's input.
        (b) FLT supplies the previous loop output as the query ($Q$) to every layer in subsequent loops; the current layer input supplies keys and values ($K,V$). The input embedding starts each FLT loop, and the first loop uses self-attention.
        Two layers and two loops are shown; each shaded column represents one forward pass. All attention is causal, and corresponding layers share parameters across loops. Normalization and within-layer residual connections are omitted for clarity.
    }
    \label{fig:main}
\end{figure}

Based on this diagnosis, we propose the \textbf{Fully Looped Transformer}, a simple redesign of the original Looped Transformer that can be trained stably across multiple loop iterations without adding learnable parameters or imposing specially designed optimization constraints. Figure~\ref{fig:main} contrasts LT and FLT. While LT passes the previous loop output only to the next loop's input, FLT combines \textbf{Fully Looped Architecture} to make it available to every layer with \textbf{Attention Injection} to control its incorporation through the existing attention module. Together, they address both the propagation path and the injection mechanism of recurrent states.

We summarize our contributions in three points:
\begin{list}{}{%
    \setlength{\leftmargin}{2.1em}%
    \setlength{\labelwidth}{1.45em}%
    \setlength{\labelsep}{0.45em}%
    \setlength{\itemsep}{2pt plus 1pt minus 1pt}%
    \setlength{\parsep}{0pt}%
    \setlength{\topsep}{0pt}%
    \setlength{\partopsep}{0pt}%
}
    \item[\textbf{·}] \textbf{Training diagnosis.} We diagnose the training instability of Looped Transformer and identify two associated phenomena: gradient oscillation and residual explosion.
    \item[\textbf{·}] \textbf{Architecture.} We introduce a parameter-free redesign that distributes recurrent signals across layers and controls their injection through existing attention, enabling stable multi-loop training without specially designed optimization constraints.
    \item[\textbf{·}] \textbf{Empirical validation.} Across scaling, ablation, and checkpoint-depth experiments, FLT remains stable through 12 loops where baselines enter high-loss regimes and improves downstream-task accuracy by 14.9\% over LT at Base size with 6 loops. Fixed-loop checkpoints improve toward their trained depth without randomized loop-count training, revealing a depth-aware test-time compute interface; extending it beyond that range is a promising direction.
\end{list}

%% file: Sections/relatedworks.tex
\section{Related Work}

\subsection{Weight-Sharing Models}

Many recurrent or looped language models can be conceptualized as weight-sharing architectures, leveraging a foundational technique to enhance parameter efficiency in deep neural networks. Universal Transformer~\citep{dehghaniuniversal} first extends this idea to the Transformer architecture by repeatedly applying the same backbone network across layers, enabling models to perform iterative refinement and endowing Transformers with the ability to simulate general computational processes. Looped Transformer~\citep{giannou2023looped} simplifies this idea by using a more general decoder-only Transformer architecture. A large body of existing work~\citep{saunshi2025reasoning, geiping2025scaling, koishekenov2025encode} has demonstrated the significant advantages of Looped Transformer in test-time scaling and reasoning tasks. Building on this line of work, we introduce a general architectural modification that requires no additional learnable parameters or computational overhead, allowing it to be integrated into existing Transformer architectures~\citep{vaswani2023attentionneed} with minimal changes.

\subsection{Training Difficulties in Recurrent Models}

Training difficulties in deep recurrent models are long-standing, with the most critical being the gradient explosion and vanishing gradient problems. \citet{bengio1994learning} systematically pointed out that when training recurrent networks using backpropagation through time (BPTT), the gradient signal decays or explodes exponentially with each time step, making it difficult for the model to learn long-range dependencies. \citet{pascanu2013difficulty} further analyzed the sources of this problem theoretically and proposed mitigation strategies such as gradient clipping. LSTM~\citep{hochreiter1997long} and GRU~\citep{cho2014learning} introduced gating mechanisms to address this issue, effectively alleviating the problem by selectively filtering and controlling the information flow. However, in the context of Looped Transformer, these challenges persist. Because the model uses a shared-weight looped structure, it is essentially equivalent to an extremely deep recurrent network, making gradient stability a core challenge in training once again.

%% file: Sections/understand.tex
\section{Diagnosing the Instability of Looped Transformer}
\label{sec:understand}

Previous studies show that increasing the number of loop iterations can substantially make LT harder to train~\citep{geiping2025scaling, zhu2025scaling}, but they do not explain the cause of such instability. To investigate this failure mode, we examine the early training dynamics of LT before full convergence. In our diagnostic runs, the relevant instability already appears during the early stage of optimization, so we use the first 2000 optimizer steps as a diagnostic window\footnote{This window is used only to identify the onset and persistence of training instability; final performance comparisons are conducted with fully trained models in Section~\ref{sec:exp}.}. In Figure~\ref{fig:understand}, we compare LT models with 6, 9, and 12 loop iterations using the same 6-layer Transformer backbone, and include a 12-loop Fully Looped Transformer (FLT) as a stable control. All models use the same pretraining setup. We monitor three quantities: training loss, residual-state norm, and gradient L2 norm of the first FFN block. Additional experiment results and training details are provided in Appendix~\ref{app:add:udr} and~\ref{app:train}.

We define training collapse as a failure mode where the model can no longer produce a usable checkpoint for evaluation. In practice, this includes entering a persistent high-loss plateau, showing severe loss fluctuations, or producing validation performance that is unusable. Under this definition, the 12-loop LT collapses in the early diagnostic run: after an initial decrease, its training loss stops improving and remains at a high plateau, while its residual-state norm continues to increase. Beyond this collapsed case, the loss curves also reveal a milder but consistent optimization difficulty for LT. As the number of loop iterations increases from 6 to 9, LT does not immediately collapse, but a clear loss gap emerges: the 9-loop LT maintains a higher training loss than the 6-loop LT throughout most of the diagnostic window. This indicates that increasing the loop count already makes optimization harder even before an outright high-loss plateau appears. In contrast, the 12-loop FLT remains stable throughout the same 2000 steps window, with smoother loss reduction, smaller residual-state norms, and more stable gradient dynamics. Based on these observations, we propose two hypotheses:

\begin{figure}[!t]
    \centering
    \includegraphics[width=\textwidth]{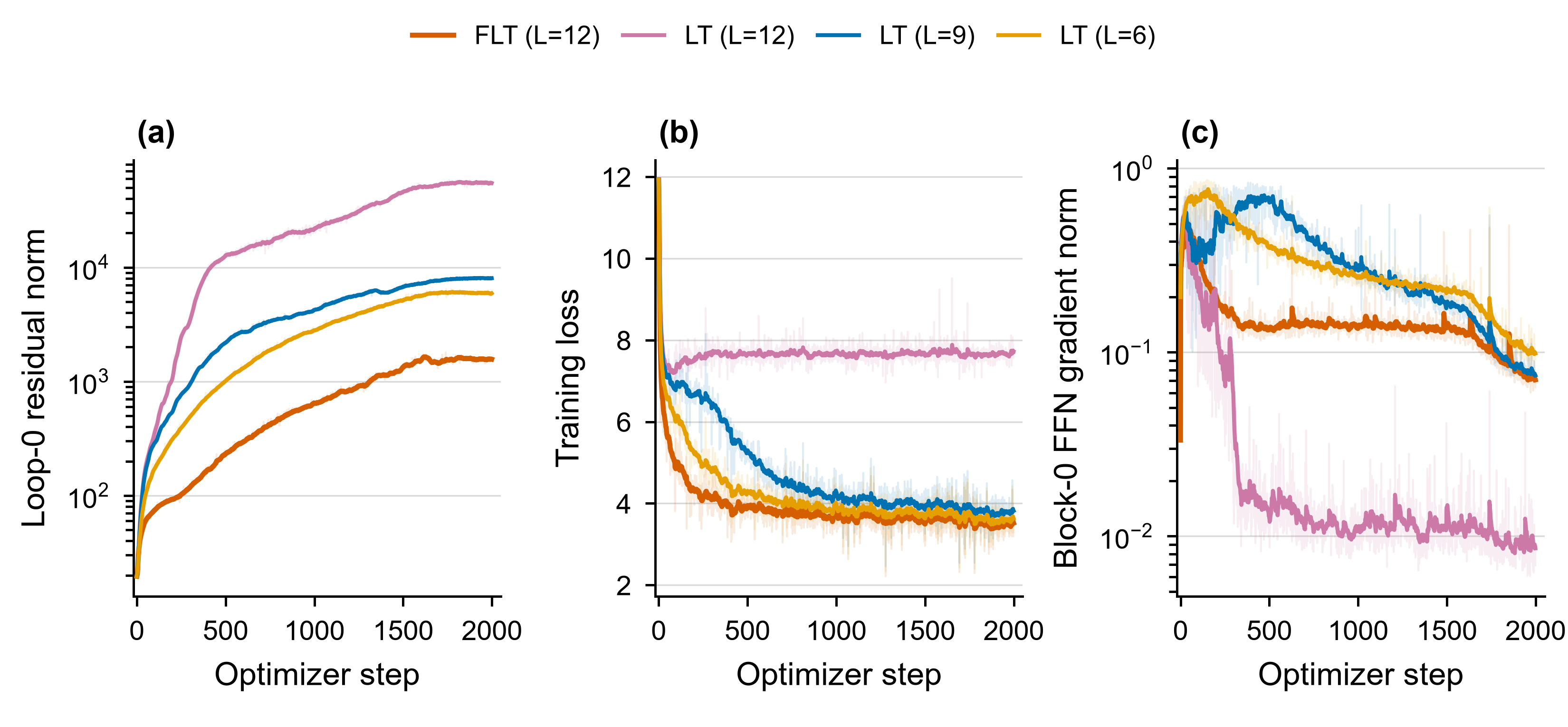}
    \caption{
    Training dynamics of LT and FLT during the first 2000 optimizer steps.
    \textbf{Left}: loop-0 residual-state norm.
    \textbf{Middle}: training loss.
    \textbf{Right}: gradient L2 norm of the first FFN block.
    Faint traces show unsmoothed values and bold curves show bias-corrected exponential moving averages with decay 0.9; the norm panels use logarithmic axes.
    }
    \label{fig:understand}
\end{figure}

\textbf{Gradient oscillation may contribute to early optimization difficulty.} Looped Transformer repeatedly applies the same Transformer backbone across loop iterations, making its training dynamics similar to those of recurrent models optimized through BPTT algorithm~\citep{werbos2002backpropagation}. Although the number of loop iterations is much smaller than the number of time steps in conventional long-sequence RNNs, gradients can still accumulate through the shared looped blocks. We therefore hypothesize that this loop-wise gradient accumulation is one possible source of early gradient oscillation, which makes optimization more sensitive and increases the likelihood of entering unstable training regimes.

\textbf{Residual explosion may become a bottleneck in highly looped settings.} Looping a fixed-depth backbone increases the effective recurrent depth of the computation graph, meaning that unrolling the looped model during training makes it equivalent to a much deeper model on the computation graph. Under this view, even a mild amplification of residual states at each loop iteration can accumulate across repeated applications of the same shared layers. We therefore hypothesize that highly looped LT can develop residual amplification in the recurrent dimension, resulting in residual explosion. Since this phenomenon persists beyond the initial transient, we treat it as a separate instability pattern that may not fully explain early collapse by itself, but can become a major optimization bottleneck when the loop count is large.

Overall, the early diagnostic results reveal two instability patterns in LT training: gradient oscillation during the early optimization phase and residual explosion at high loop counts. These observations motivate the design choices introduced in the next section, where we aim to stabilize looped training without adding learnable parameters.

%% file: Sections/architecture.tex
\section{Fully Looped Transformer}
\label{sec:arch}

In Section~\ref{sec:understand}, we identified two patterns associated with Looped Transformer's training instability: gradient oscillation and residual explosion. We design Fully Looped Transformer to mitigate these two problems. FLT introduces two improvements: (1) Fully Looped Architecture (FLA) and (2) Attention Injection (AI). Figure~\ref{fig:main} shows a comparison of the overall architectures between FLT and LT, as well as the specific implementations of FLA and AI in FLT.

\subsection{Fully Looped Architecture}

Although all of our models already incorporate RMSNorm~\citep{zhang2019rootmeansquarelayer} in each layer to regulate the output residual states, residual explosion still occurs. This suggests that normalization alone is insufficient, and that the issue must be addressed from a deeper structural perspective. We use $D$ for the number of unique Transformer layers, $L$ for the number of training loops, and $K$ for the number of inference loops.

In Section~\ref{sec:understand}, we suggest that residual explosion may be due to a large effective recurrent depth of the model, which poses similar
optimization challenges as extremely deep neural networks. Extremely deep neural networks require shortcut connections like residual connections~\citep{he2016deep} to alleviate their optimization difficulties. \citet{veit2016residual} indicate that residual connections can help optimize deep models because they are equivalent to reducing the depth of the model during the optimization process. Inspired by these studies, we aim to design a residual connections that works in the recurrent dimension to mitigate the optimization difficulties related to effective recurrent depth.

We propose \textbf{Fully Looped Architecture (FLA)} as a recurrent connectivity pattern. In vanilla Looped Transformer, the output of the previous loop iteration is passed only to the first layer of the next iteration. As a result, layers at larger depth can access the recurrent state only through a long chain of intermediate transformations. FLA changes this by making the previous loop output available to every layer in the current iteration. Formally, let $h_D^{(t-1)}$ denote the output hidden state of the previous loop iteration. For each layer $l$ in the current iteration $t$, FLA defines the layer update as:

\begin{equation}
    \mathbf{h}_l^{(t)} = f_\theta^{(l)}\!\left(\mathbf{h}_{l-1}^{(t)},\,
    \mathbf{h}_D^{(t-1)}\right), \quad l = 1, \dots, D,
\end{equation}

Here, $f_\theta^{(l)}$ denotes the $l$-th Transformer layer equipped with a fusion operation that incorporates $h_D^{(t-1)}$. FLA itself only specifies the recurrent connectivity; it does not prescribe how the previous loop state should be fused into each layer. In our ablations, we consider a direct residual-addition implementation, denoted FLT$_{res}$. In the complete Fully Looped Transformer, we implement this fusion using Attention Injection, which is described in the next subsection.

\subsection{Attention Injection}

In Section~\ref{sec:understand}, we observed abnormal gradient dynamics during the early stage of pretraining. These dynamics are consistent with the difficulty of RNNs trained with the BPTT algorithm. Gradient clipping~\citep{pascanu2013difficulty} is a common practical mitigation, but it operates as an optimization-level intervention and introduces an additional threshold hyperparameter. Here, we explore an architectural mechanism that controls the recurrent signal.

LSTM~\citep{hochreiter1997long} and GRU~\citep{cho2014learning} mitigate the gradient explosion problem through gating mechanisms. A key role of these gates is to modulate the information in residual flow within a bounded range, which helps control the magnitude of propagated signals and gradients. Inspired by this principle, we propose \textbf{Attention Injection (AI)}, which reuses the Self-Attention module as a Cross-Attention module to inject the previous loop iteration's hidden state into the current loop step. Instead of adding $h_D^{(t-1)}$ directly into the residual flow, AI uses it as the query of an attention operation. The current layer representation provides the keys and values. In this way, the previous loop state determines which information to retrieve, but the injected activation is constructed from the current value vectors rather than copied directly from the previous residual state.

By routing $\mathbf{h}_D^{(t-1)}$ as the query $Q$, AI can be viewed as using Softmax operations to control and select information from the residual flow. When $h^{(t-1)}_D$ is used as the query, it influences the injected signal indirectly through attention blocks rather than being directly added to the residual flow. Because the attention weights are normalized by the softmax operation, the injected signal is a normalized mixture of the value vectors. Therefore, the magnitude of the injected recurrent signal is mediated by the current value stream rather than being directly determined by $h_D^{(t-1)}$. This provides a controlled mixing mechanism for recurrent information and reduces the risk of directly amplifying the residual flow. Formally, the standard attention operation is:
\begin{equation}
    \text{Attention}(Q, K, V) = \text{Softmax}\!\left(\frac{QK^\top}
    {\sqrt{d_k}}\right) V,
\end{equation}

Concretely, during the first loop iteration ($t=1$), the model performs standard Self-Attention. In subsequent loop iterations ($t > 1$), the model switches to Cross-Attention: the hidden state from the previous loop iteration $\mathbf{h}_D^{(t-1)}$ is used as the query ($Q$), while the output of the preceding module $\mathbf{z}_l^{(t)}$ serves as both the key ($K$) and value ($V$). Formally, the attention block for $t>1$ is modified as:
\begin{equation}
    \mathbf{a}_l^{(t)} =\mathrm{Attention}\left(Q=W_Q h_D^{(t-1)}, K=W_K z_l^{(t)}, V=W_V z_l^{(t)} \right),
\end{equation}
where $\mathbf{a}_l^{(t)}$ denotes the output of the attention block at layer $l$ in loop iteration $t$. In this design, the previous residual state is not directly added into the residual flow. Instead, it acts as a query that selects and reuses information from the current layer representation.

Inspired by Input Injection~\citep{anil2022pat}, in the first layer of the model, AI directly uses the input embedding $\mathbf{x}$ as the $z_l^{(t)}$. Input Injection adds the input embedding $\mathbf{x}$ to the residual flow at each loop iteration as a method of emphasizing the input signal. Input Injection has been used as an empirical stabilization technique in prior looped models~\citep{zhu2025scaling}. In our architecture, we hypothesize that first-layer Attention Injection serves a role analogous to Input Injection by reinforcing the input signal through the attention mechanism.

AI reuses the Self-Attention projection matrices $W_Q,W_K,W_V$ as Cross-Attention projection matrices. No additional Cross-Attention parameters, gates, or normalization layers are introduced. We deliberately inject the recurrent signal through $Q$ rather than $K$ or $V$. This design keeps the key-value streams in the same form as standard attention, preserving compatibility with existing KV-cache reuse mechanisms during loop iteration.

\subsection{Implementation Details}

In this work, We use the pretrained tokenizer of Qwen3~\citep{yang2025qwen3technicalreport} throughout training. We used the $\mu$P~\citep{yang2022tensorprogramsvtuning} method to adjust the learning rate of models with different depths and optimized the models using the Muon~\citep{jordan2024muon} and AdamW~\citep{loshchilov2019decoupledweightdecayregularization} optimizers. A fixed batch size was used for all models, and the total number of training tokens was scaled according to the Chinchilla scaling law~\citep{hoffmann2022training} to control the total number of training steps. When comparing two models with an equal number of parameters, we use exactly the same hyperparameters. Following \citet{geiping2025scaling}, we use the standard next-token prediction loss, supervising only the last step. Our Transformer backbone network references the implementation of \citet{nanochat}. For more implementation details, please refer to the Appendix~\ref{app:train} and~\ref{app:arc}.

%% file: Sections/experiment.tex
\section{Experiments}
\label{sec:exp}

\subsection{Experimental Setup}
\label{sec:exp:setup}

\textbf{Dataset.} We used FineWeb-Edu~\citep{lozhkov2024fineweb-edu} as our pretraining dataset and validation dataset. FineWeb-Edu is a widely used high-quality pretraining dataset containing 200B of high-value tokens. The dataset itself has already been deduplicated and refined.
Following the Chinchilla scaling law~\citep{hoffmann2022training}, we scaled the number of training tokens used proportionally to the total number of model parameters, using 20 tokens per parameter.

\begin{figure}[!t]
    \centering
    \includegraphics[width=0.92\textwidth]{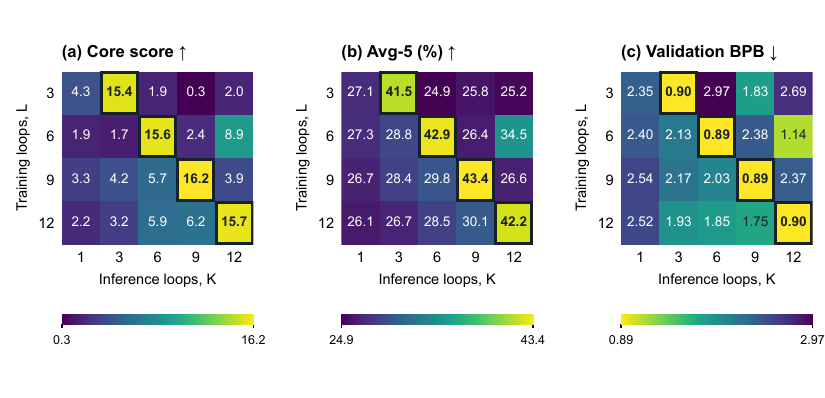}
    \caption{
     \textbf{Test-time loop adaptation of Base-size FLT.}
     Panels show Core, Avg$_5$, and validation BPB; rows are checkpoints, and outlines mark $K=L$.
    }
    \label{fig:tts}
\end{figure}

\textbf{Evaluation.} In Table~\ref{tab:main} and Figure~\ref{fig:tts}, we present WikiText-2 perplexity (PPL)~\citep{merity2016pointer}, validation bits per byte (BPB), the Core Metric, and accuracy on five downstream benchmarks. Core, proposed by DCLM~\citep{li2024datacomp}, aggregates 21 benchmark task families for evaluating language models during pretraining. More details, including the centered-accuracy definition and benchmark configurations, are provided in Appendix~\ref{app:exp}.

\textbf{Baseline.} In the Loop Scaling Experiments, we compare the Fully Looped Transformer (FLT) against three baselines: (1) the original Looped Transformer (LT); (2) the Looped Transformer with Input Injection (LT$_{i}$); and (3) the Looped Transformer with Attention Injection applied only at the first layer (LT$_{ai}$). All baselines are trained at two scales: the \textbf{small} size (127M parameters, 6 layers) and the \textbf{base} size (318M parameters, 12 layers). When the parameter counts and loop iterations are the same, all models use the same amount of computation during pretraining.

\textbf{Ablation.} In the ablation experiment, We check the stability of different methods using the same settings as Section~\ref{sec:understand}. In addition to the baselines mentioned above, we also compared the Fully Looped Transformer variant (FLT$_{res}$), which implements the Fully Looped Architecture by directly adding the hidden state to the residual flow instead of using Attention Injection. This variant exhibited training collapse under various looping settings, therefore, we do not directly compare it with other architectures in the loop scaling experiment, but instead to compare the characteristics of different architectures in the ablation experiment. By default, we implement Attention Injection with Full Attention (FA)~\citep{vaswani2023attentionneed}. We also experimented to see if Attention Injection could be applied to different attention mechanism variants, including Sliding Window Attention (SWA)~\citep{beltagy2020longformerlongdocumentTransformer}, Multi-head Latent Attention (MLA)~\citep{deepseekai2024deepseekv2strongeconomicalefficient} and Grouped-Query Attention (GQA)~\citep{ainslie2023gqatraininggeneralizedmultiquery}. Please refer to Appendix~\ref{app:train} and~\ref{app:arc} for more experiment and training setup details.

\input{Tables/table1}

\subsection{Loop Scaling Experiments}

As shown in Table~\ref{tab:main}, the original LT degrades as the number of loop iterations increases. At the small size, its downstream-task average drops from 33.69 to 30.92, while Wiki2 PPL, validation BPB, and the Core Metric also worsen. LT$_i$ remains stable, reaching 34.15 at 6 loops, while LT$_{ai}$ is severely degraded in this setting but recovers at 9 loops, reaching 34.28. FLT trains stably across all loop settings and improves from 33.40 to 33.83. These results suggest that the modified variants are less sensitive to increased loop depth than the original LT.

At the Base size, FLT is the only evaluated variant whose average downstream accuracy increases monotonically across 3, 6, and 9 loops. The original LT drops from 41.71 to 37.32 between 3 and 6 loops and collapses at 9 loops, while LT$_i$ and LT$_{ai}$ show only fluctuating or saturated gains. At 9 loops, FLT reaches 43.43, achieving the best average score, the lowest Wiki2 PPL, the highest Core Metric, and a tied-best validation BPB. FLT thus benefits more reliably from additional loops.

The two severe degradations in Table~\ref{tab:main} are not missing evaluations. Base LT at 9 loops is clearest: PPL rises to 3871.50 and Core falls below zero after worsening from 3 to 6 loops, suggesting that repeated unrolling pushes optimization toward instability. Small LT$_{ai}$ is non-monotonic: its degraded 6-loop run recovers to an Avg$_5$ of 34.28 at 9 loops when trained separately, indicating configuration sensitivity rather than a deterministic threshold.

Overall, FLT provides the most reliable base-size scaling: from 3 to 9 loops, PPL falls 40.48$\rightarrow$39.44, Core rises 15.38$\rightarrow$16.17, and BPB improves 0.898$\rightarrow$0.893. Avg$_5$ exceeds LT by 5.57 points at 6 loops and reaches 43.43 at 9 loops, showing consistent gains from added recurrent computation and avoiding comparable degradation in the tested settings.

\begin{figure}[!t]
    \centering
    \includegraphics[width=\textwidth]{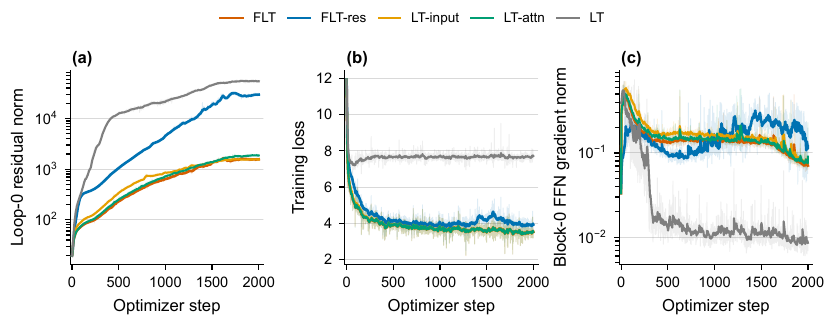}
    \caption{
    Training dynamics of FLT and LT variants with matched parameter and loop counts.
    \textbf{Left}: loop-0 residual-state norm; \textbf{middle}: training loss; \textbf{right}: first-FFN gradient L2 norm.
    Faint traces are raw values; bold curves are bias-corrected EMA (decay 0.9). Norm axes are logarithmic.
    }
    \label{fig:abl}
\end{figure}

\begin{figure}[!t]
    \centering
    \includegraphics[width=\linewidth]{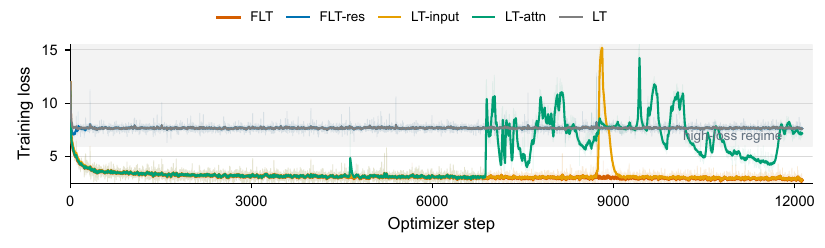}
    \caption{Training loss at 12 loops and 12 layers. FLT optimizes at low loss; LT and FLT-res remain at high loss, while injection variants exhibit transient instabilities. Faint traces: raw values; bold curves: bias-corrected EMA (decay 0.9).}
    \label{fig:left}
\end{figure}

\subsection{Ablation Experiments}

Figures~\ref{fig:abl} and~\ref{fig:left}, together with Table~\ref{tab:right}, compare the two proposed components through training dynamics and final evaluation, highlighting how they jointly improve stability.

\textbf{Fully Looped Architecture attenuates residual explosion.} FLT$_{res}$ substantially reduces residual-state norms relative to LT, although residual explosion persists. Combining FLA with Attention Injection further reduces residual amplification, with no residual explosion observed. Thus, FLA attenuates residual amplification but does not alone guarantee stable training.

\textbf{First-layer Attention Injection exhibits effects similar to Input Injection.} When applied only at the first layer, Attention Injection achieves performance comparable to Input Injection in several settings, suggesting that using the input embedding as the key and value stream can preserve part of the effect of direct input injection. This similarity is limited, however, as shown by the severe degradation of Small LT$_{ai}$ at 6 loops.

\noindent
\begin{minipage}[t]{0.48\textwidth}
\vspace{0pt}
\textbf{Compatibility with attention variants.} The Base-size, 12-loop FLT variants in Table~\ref{tab:right} complete training stably with performance comparable to Full Attention, supporting compatibility across attention mechanisms. The comparable scores and stable training across variants suggest that the benefit arises from the Attention Injection interface rather than from any particular attention kernel. Appendix~\ref{app:add:abl} provides further results.
\end{minipage}\hfill
\begin{minipage}[t]{0.48\textwidth}
\vspace{0pt}
\centering
\setlength{\tabcolsep}{4pt}
\renewcommand{\arraystretch}{1.2}
\captionsetup{position=top,skip=6pt}
\captionof{table}{Base-size FLT with 12 loops and different attention variants. All four variants train stably with comparable evaluation scores.}
\label{tab:right}
\begin{tabular}{@{}l|rrrr@{}}
    \hline
    \textbf{Metrics} & \textbf{GQA} & \textbf{MLA} & \textbf{SWA} & \textbf{FA} \\
    \hline
    PPL$\downarrow$ & 39.68 & 38.76 & 38.91 & 41.12 \\
    BPB$\downarrow$ & 0.897 & 0.904 & 0.895 & 0.895 \\
    Core$\uparrow$ & 16.24 & 15.64 & 15.58 & 15.66 \\
    \hline
\end{tabular}
\end{minipage}

\textbf{FLA and Attention Injection jointly provide the most stable training in our comparisons.} Input Injection and first-layer Attention Injection can remain unstable at 12 loops (Figure~\ref{fig:left}); FLA alone also exhibits unstable gradients (Figure~\ref{fig:abl}).

\par

\subsection{Test-Time Compute Adaptation Experiments}

We use fixed-loop FLT checkpoints to probe how recurrent depth is learned at inference time. Across Core, downstream-task accuracy, and validation BPB, every checkpoint performs best at its training depth, $K=L$. Performance generally improves as the inference depth approaches the trained depth for the 9- and 12-loop checkpoints, showing that fixed-loop training can make additional inference computation useful even without randomized loop-count training. However, matched-depth performance improves through 9 loops and then declines at 12 loops, indicating that deeper training is not automatically better.

Beyond the trained depth, the checkpoints behave differently: the 9-loop checkpoint suffers a severe drop at 12 loops, while the 6-loop checkpoint partially recovers but remains below its matched-depth result. These off-diagonal patterns suggest that FLT learns depth-specialized recurrent computation rather than a universal rule for arbitrary-depth extrapolation. Thus, FLT provides a useful depth-aware test-time compute interface within the trained regime, while reliable extrapolation beyond that regime remains a promising direction for future work.

%% file: Tables/table1.tex
\begin{table}[t]
\centering
\small
\setlength{\tabcolsep}{2.5pt}
\renewcommand{\arraystretch}{1.08}
\caption{\textbf{Language modeling and downstream evaluation across loop counts.} Small and Base use 6 and 12 layers, respectively. PPL is measured on WikiText-2; BPB is validation bits per byte. Avg$_5$ is the unweighted mean of the five displayed downstream tasks, computed before rounding. All evaluated configurations, including severely degraded models, are reported. Bold denotes column-best values across all rows, with ties at the displayed precision.}
\label{tab:main}
\begin{tabular*}{\textwidth}{@{}@{\extracolsep{\fill}} l | l c | ccc | ccccc | c @{}}
\toprule
\multicolumn{3}{c}{\textbf{Configuration}} & \multicolumn{3}{c}{\textbf{LM / Core}} & \multicolumn{5}{c}{\textbf{Downstream accuracy (\%)}} & \textbf{Mean} \\
\cline{1-3}\cline{4-6}\cline{7-11}\cline{12-12}
Model & Size & Loops & PPL$\downarrow$ & BPB$\downarrow$ & Core$\uparrow$ & LMB$\uparrow$ & PIQA$\uparrow$ & Hella$\uparrow$ & ARC-E$\uparrow$ & ARC-C$\uparrow$ & Avg$_5\uparrow$ \\
\midrule
\multirow{6}{*}{\textbf{LT}} & \multirow{3}{*}{Small} & 3 & 67.80 & 1.055 & 8.14 & 18.71 & 59.47 & 27.66 & 38.97 & 23.63 & 33.69 \\
 &  & 6 & 73.04 & 1.068 & 7.20 & 16.34 & 57.56 & 26.80 & 36.66 & 23.29 & 32.13 \\
 &  & 9 & 78.79 & 1.115 & 4.67 & 12.32 & 56.91 & 26.00 & 36.49 & 22.87 & 30.92 \\
\cline{2-12}
 & \multirow{3}{*}{Base} & 3 & 43.13 & 0.905 & 16.14 & 30.08 & 64.25 & 35.43 & 51.64 & 27.13 & 41.71 \\
 &  & 6 & 51.39 & 0.948 & 11.44 & 26.37 & 61.32 & 31.37 & 43.22 & 24.32 & 37.32 \\
 &  & 9 & 3871.50 & 2.373 & -2.19 & 0.00 & 49.24 & 24.50 & 28.16 & 22.78 & 24.93 \\
\midrule
\multirow{6}{*}{\textbf{LT$_i$}} & \multirow{3}{*}{Small} & 3 & 62.76 & 1.044 & 6.99 & 20.05 & 59.63 & 27.80 & 39.56 & 22.44 & 33.90 \\
 &  & 6 & 62.54 & 1.042 & 7.34 & 20.40 & 59.25 & 27.71 & 39.94 & 23.46 & 34.15 \\
 &  & 9 & 62.31 & 1.045 & 8.38 & 20.59 & 58.65 & 27.76 & 38.85 & 24.23 & 34.02 \\
\cline{2-12}
 & \multirow{3}{*}{Base} & 3 & 40.46 & 0.901 & 15.36 & 31.13 & 64.69 & 35.80 & \textbf{56.06} & \textbf{28.41} & 43.22 \\
 &  & 6 & 39.59 & 0.900 & 15.30 & 29.92 & 64.64 & 36.34 & 52.65 & 26.28 & 41.97 \\
 &  & 9 & 43.43 & 0.901 & 14.73 & 30.56 & 64.85 & 36.25 & 53.79 & 27.82 & 42.65 \\
\midrule
\multirow{6}{*}{\textbf{LT$_{ai}$}} & \multirow{3}{*}{Small} & 3 & 66.12 & 1.047 & 7.18 & 19.76 & 58.87 & 27.50 & 39.60 & 23.55 & 33.86 \\
 &  & 6 & 5137.10 & 2.199 & 2.99 & 0.12 & 53.43 & 25.08 & 30.22 & 23.81 & 26.53 \\
 &  & 9 & 73.59 & 1.047 & 6.47 & 19.66 & 59.85 & 27.56 & 39.69 & 24.66 & 34.28 \\
\cline{2-12}
 & \multirow{3}{*}{Base} & 3 & 41.64 & 0.901 & 14.10 & 29.54 & 62.95 & 35.82 & 51.98 & 27.65 & 41.59 \\
 &  & 6 & 41.49 & 0.900 & 14.98 & 30.35 & 62.89 & 36.38 & 52.82 & 26.62 & 41.81 \\
 &  & 9 & 42.44 & 0.902 & 13.75 & 28.39 & 64.42 & 35.95 & 50.97 & 28.07 & 41.56 \\
\midrule
\multirow{6}{*}{\textbf{FLT}} & \multirow{3}{*}{Small} & 3 & 62.77 & 1.045 & 5.72 & 18.40 & 59.14 & 27.48 & 39.56 & 22.44 & 33.40 \\
 &  & 6 & 60.08 & 1.039 & 6.35 & 19.25 & 58.98 & 27.91 & 37.46 & 23.04 & 33.33 \\
 &  & 9 & 63.30 & 1.040 & 6.94 & 19.04 & 59.14 & 28.20 & 38.89 & 23.89 & 33.83 \\
\cline{2-12}
 & \multirow{3}{*}{Base} & 3 & 40.48 & 0.898 & 15.38 & 30.64 & 63.82 & 35.73 & 51.09 & 26.37 & 41.53 \\
 &  & 6 & 39.51 & \textbf{0.893} & 15.61 & 31.52 & 64.85 & 37.03 & 52.82 & 28.24 & 42.89 \\
 &  & 9 & \textbf{39.44} & \textbf{0.893} & \textbf{16.17} & \textbf{32.76} & \textbf{65.07} & \textbf{37.05} & 53.87 & \textbf{28.41} & \textbf{43.43} \\
\bottomrule
\end{tabular*}
\end{table}

%% file: Sections/conclusion.tex
\section{Conclusion}

High-depth Looped Transformer training suffers from gradient oscillation and residual explosion. Our parameter-free FLT exposes recurrent states to every layer and injects them through attention, enabling stable 12-loop training. Across loop-scaling, ablation, attention-variant, and test-time compute adaptation experiments, FLT shows downstream gains, attention compatibility, and depth-aware inference, offering a promising foundation for future test-time scaling.

%% file: Sections/appendix.tex
\newpage
\section{Appendix}
\label{appendix}

\subsection{Additional Experiments Results}
\label{app:add}

\subsubsection{Diagnostic Experiment}
\label{app:add:udr}

\begin{figure}[!h]
    \centering
    \includegraphics[width=\linewidth]{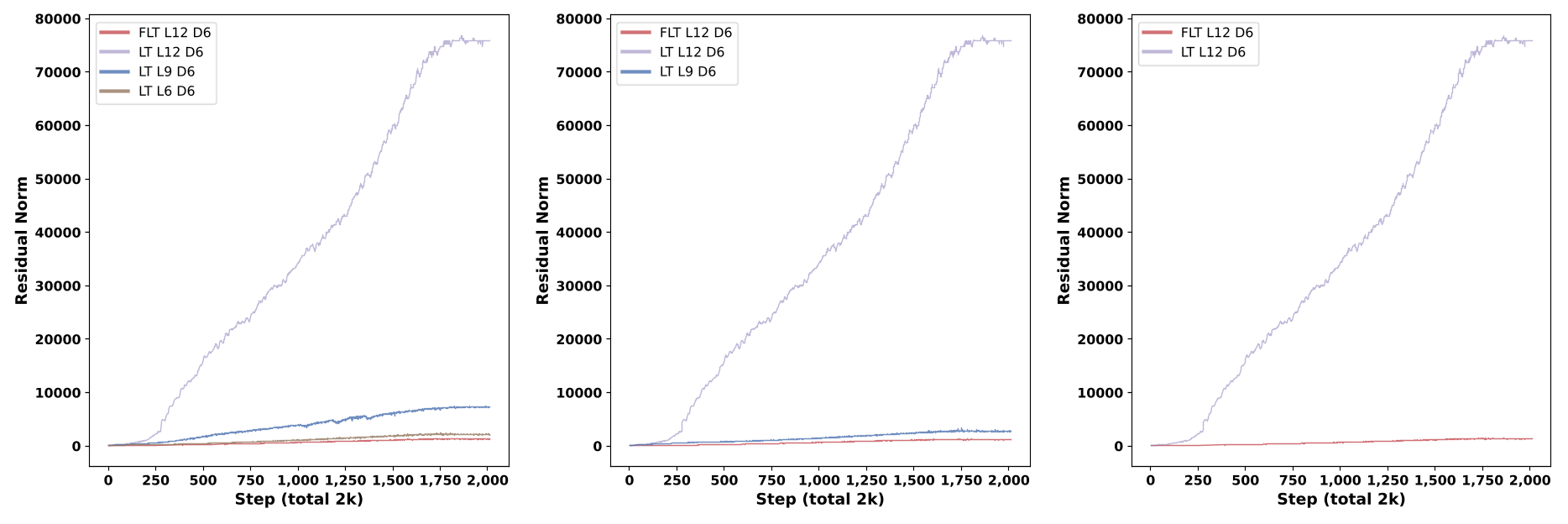}
    \caption{
    \textbf{Left}: the residual norm at the 6th loop iteration.
    \textbf{Middle}: the residual norm at the 9th loop iteration.
    \textbf{Right}: the residual norm at the 12th loop iteration.
    }
    \label{fig:loops}
\end{figure}

\begin{figure}[!h]
    \centering
    \includegraphics[width=\linewidth]{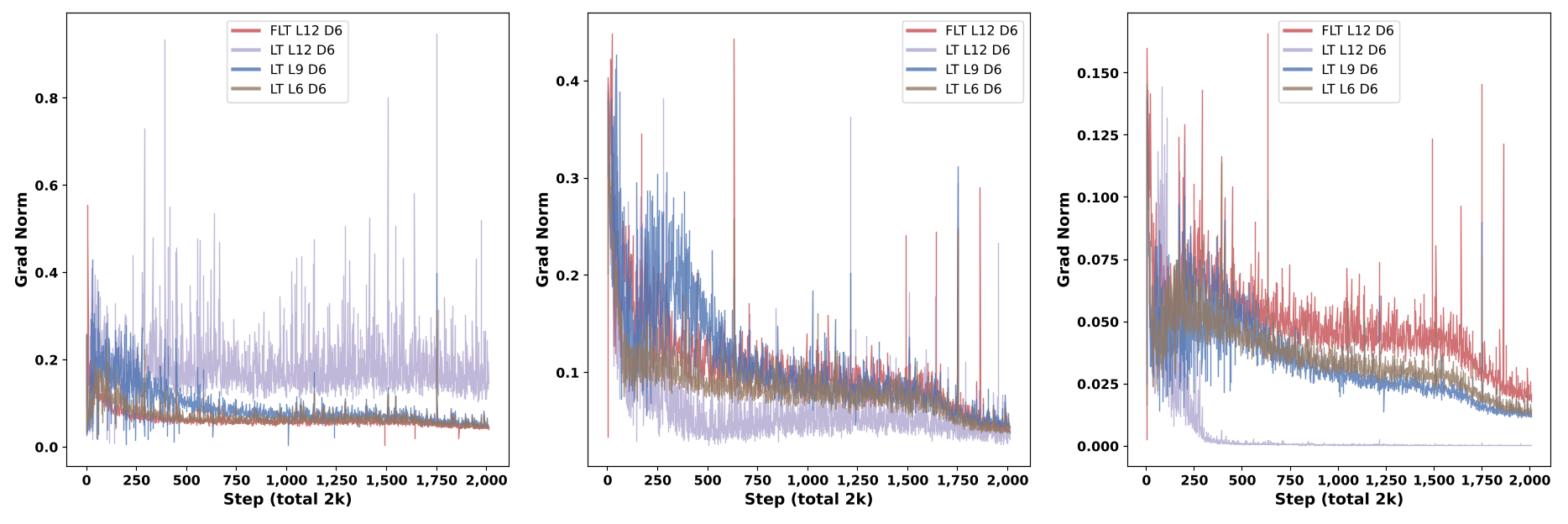}
    \caption{
    \textbf{Left}: the gradient norm of LM head block.
    \textbf{Middle}: the gradient norm of FFN at 5th layer.
    \textbf{Right}: the gradient norm of attention block at 5th layer.
    }
    \label{fig:grads}
\end{figure}

Figures~\ref{fig:loops} and~\ref{fig:grads} provide supplementary evidence for the diagnostic experiment in Section~\ref{sec:understand}. Figure~\ref{fig:loops} extends the residual-state norm analysis to later loop iterations. Compared with the first-loop residual norm shown in the main text, these results show that the residual amplification of vanilla LT becomes more pronounced as the recurrent computation proceeds, especially under larger loop counts. In contrast, FLT keeps the residual-state norms substantially smaller and more stable across loop iterations. This further supports our diagnosis that highly looped LT suffers from residual explosion, while FLT mitigates this instability by stabilizing recurrent information propagation.

Figure~\ref{fig:grads} reports gradient norms from additional modules, including the LM head, an intermediate FFN block, and an intermediate attention block. These results show that the gradient oscillation observed in the main diagnostic figure is not restricted to the first FFN block, but also appears in other parts of the model during early training. Therefore, Figures~\ref{fig:loops} and~\ref{fig:grads} together strengthen the conclusion that LT instability is associated with both residual explosion and gradient oscillation, whereas FLT alleviates both phenomena.

\subsubsection{Ablation Experiment}
\label{app:add:abl}

\begin{figure}[!h]
    \centering
    \includegraphics[width=0.6\linewidth]{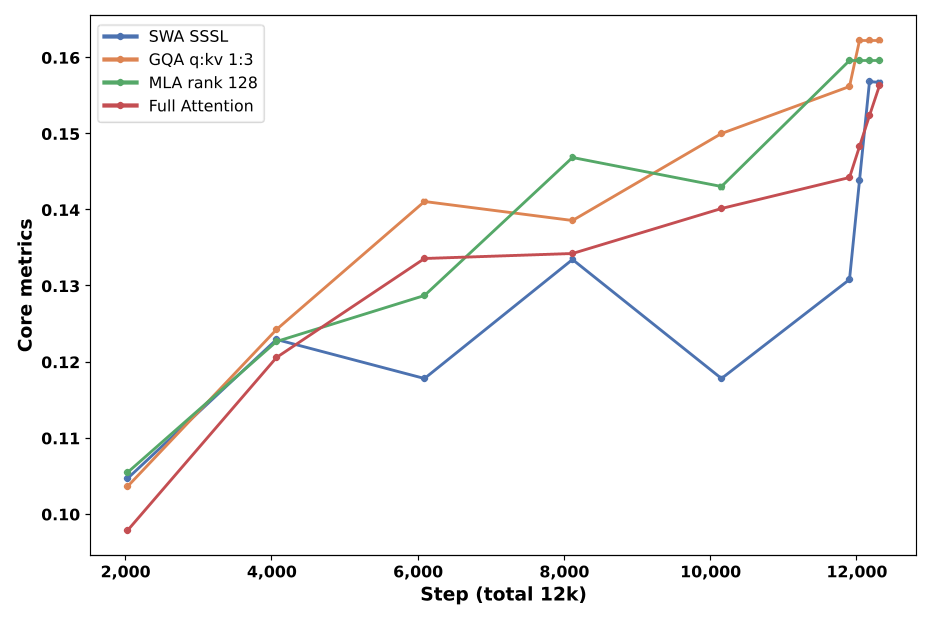}
    \caption{
    Trend chart of Core Metric changes for FLT with different attention variants throughout the training process
    }
    \label{fig:trend}
\end{figure}

\begin{figure}[!h]
    \centering
    \includegraphics[width=0.8\linewidth]{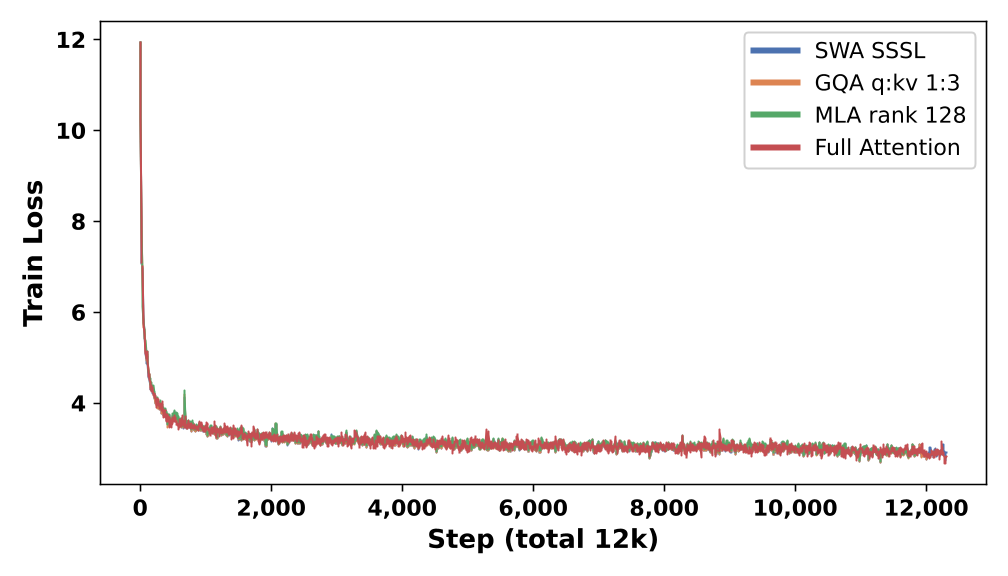}
    \caption{The training loss of FLT with different attention variants. Smoothed with factor 0.9.}
    \label{fig:loss}
\end{figure}

Figures~\ref{fig:trend} and~\ref{fig:loss} provide supplementary results for the ablation experiment in the main text. Figure~\ref{fig:trend} shows the Core Metric trajectories of FLT when Attention Injection is implemented with different attention variants, including Full Attention, Sliding Window Attention, Multi-head Latent Attention, and Grouped-Query Attention. The comparable trends across these variants indicate that the effectiveness of Attention Injection is not tied to a single attention implementation.

Figure~\ref{fig:loss} further reports the corresponding training-loss curves. All attention variants train stably without collapse, which is consistent with the results in Table~\ref{tab:right}. These observations support the conclusion that Attention Injection is compatible with multiple attention mechanisms and can serve as a robust stabilization component for Fully Looped Transformer.

\subsubsection{Failure Cases in Loop Scaling}
\label{app:add:failure}

\textbf{Severe degradation rather than missing evaluations.} Table~\ref{tab:main} reports two severely degraded configurations: Base LT with 9 loops and Small LT$_{ai}$ with 6 loops. Their WikiText-2 PPL reaches 3871.50 and 5137.10, validation BPB reaches 2.373 and 2.199, and LAMBADA accuracy falls to 0.00\% and 0.12\%, respectively. The joint deterioration of language-modeling and downstream metrics distinguishes these cases from modest performance fluctuations. Here, failure refers to severely degraded predictive performance, not an inability to execute evaluation or a failed evaluation job; the measured scores are retained in Table~\ref{tab:main}.

\textbf{Base LT deteriorates before the 9-loop failure.} As the loop count increases from 3 to 6 to 9, PPL worsens from 43.13 to 51.39 to 3871.50, while Core decreases from 16.14 to 11.44 to $-2.19$. Thus, severe degradation is preceded by declining performance at a lower loop count. This pattern is consistent with the optimization difficulties discussed in Section~\ref{sec:understand}: repeated application of shared blocks lengthens the unrolled computation path, along which residual amplification and unstable gradient propagation may hinder learning. However, the diagnostic curves in Figures~\ref{fig:understand}--\ref{fig:left} use different configurations and do not establish the cause of this particular failure.

\textbf{Small LT$_{ai}$ shows non-monotonic sensitivity to loop count.} Although the 6-loop configuration is severely degraded, the separately trained 9-loop model attains PPL 73.59 and Avg$_5$ 34.28. These are independent training configurations, not successive stages of one training trajectory. Changing the loop count changes both the forward computation and its gradient paths, so failure at one count need not imply failure at every larger count. Unlike FLT, LT$_{ai}$ applies Attention Injection only at the first layer and lacks the full layer-wise recurrent connectivity. Its results show that first-layer injection alone does not ensure reliable optimization across the tested settings.

Together, these cases suggest configuration-dependent optimization sensitivity rather than a deterministic loop-count threshold. Initialization, data order, and hyperparameter sensitivity are possible contributors, but the reported runs do not isolate their effects or establish failure probabilities across seeds. FLT avoids similarly severe degradation in all configurations in Table~\ref{tab:main}; this supports its robustness under the tested protocol, not a guarantee of stability under arbitrary training conditions.

\subsection{Training Details}
\label{app:train}

\subsubsection*{Data}

We train on FineWeb-Edu~\citep{lozhkov2024fineweb-edu}, streamed from Parquet files
and tokenized on-the-fly with the pretrained Qwen3 BPE tokenizer. The tokenizer has vocabulary size
$|\mathcal{V}| = 151643$ (padded to $151644$ for divisibility).
Data is loaded in a DDP-aware manner: each rank reads disjoint row groups from
the Parquet shards. The final Parquet shard is held out as the validation set.

\subsubsection*{Optimization}

We use two optimizers running in parallel, following the approach of modded-nanogpt~\citep{modded_nanogpt_2024}:

\paragraph{Muon~\citep{jordan2024muon} for transformer block linear layers.}
Muon runs standard SGD with momentum internally, then replaces each 2-D
parameter update $G$ with the nearest orthogonal matrix via a 5-step
Newton--Schulz iteration:
\begin{equation}
  X \leftarrow aX + b(XX^\top)X + c(XX^\top)^2 X,
  \quad (a,b,c) = (3.4445,\,-4.7750,\,2.0315).
\end{equation}
We use learning rate $\eta_{\mathrm{Muon}} = 0.02$ and momentum
$\mu = 0.95$ (warmed up from $0.85$ over the first $300$ steps).

\paragraph{AdamW~\citep{loshchilov2019decoupledweightdecayregularization} for embeddings and LM head.}
We use $\beta_1 = 0.8$, $\beta_2 = 0.95$, $\varepsilon = 10^{-10}$,
no weight decay.
Learning rates are set to $\eta_{\mathrm{embed}} = 0.2$ and
$\eta_{\mathrm{head}} = 0.004$, and are scaled by
$\sqrt{768 / d_{\mathrm{model}}}$ for models with $d_{\mathrm{model}} \neq 768$,
following the $\mu$P-style depth-scaling convention.

\subsubsection*{Learning Rate Schedule}

We use a trapezoidal schedule with no warm-up:
the learning rate is held constant for the first $80\%$ of training, then
linearly decayed to zero over the final $20\%$.
All parameters share the same multiplicative schedule factor.

\subsubsection*{Batch Size and Gradient Accumulation}

We use a fixed total batch size of $2^{19} = 524{,}288$ tokens per optimizer
step, with a sequence length of $T = 1024$.
When training on $N_{\mathrm{GPU}}$ GPUs each processing a per-device batch of
$B_{\mathrm{dev}}$ sequences, gradient accumulation runs for
$\lceil 524{,}288 / (B_{\mathrm{dev}} \cdot T \cdot N_{\mathrm{GPU}}) \rceil$
micro-steps automatically.

\subsubsection*{Training Length}

Unless otherwise specified, we follow the Chinchilla optimal
ratio~\cite{hoffmann2022training} and train for a number of steps such that
the total token count equals $20 \times N_{\mathrm{params}}$, where
$N_{\mathrm{params}}$ denotes the total number of model parameters.

\subsubsection*{Regularization and Precision}

We apply gradient clipping with maximum norm $1.0$.
No dropout or weight decay is used. All forward passes and backward passes use bfloat16 mixed precision (\texttt{torch.amp.autocast}); the token embedding is stored in bfloat16 throughout training. \texttt{torch.compile} (TorchInductor, \texttt{dynamic=False}) is enabled for the training graph.

\subsubsection*{Distributed Training}

We use PyTorch DDP with \texttt{torchrun}.
The Muon optimizer uses a distributed variant (DistMuon) where Newton--Schulz
orthogonalization is computed independently on each rank but momentum buffers
are kept in sync. The AdamW optimizer uses ZeRO-2~\citep{rajbhandari2020zeromemoryoptimizationstraining} sharding (DistAdamW) to reduce optimizer state memory. Our models were trained on 8$\times$ A100 GPUs with 80GB RAM and 32$\times$ Intel Xeon CPU cores. Each model was trained for at most five days.

\subsubsection*{Diagnostic metrics}

For the diagnostic experiments in Section~\ref{sec:understand}, we record the following quantities every evaluation interval.
The residual-state norm is the mean L2 norm of the loop-output hidden state,
averaged over the batch and sequence dimensions:
\[
\mathrm{ResNorm}^{(t)} =
\frac{1}{B T}\sum_{b=1}^{B}\sum_{i=1}^{T}
\left\| h^{(t)}_{D,b,i} \right\|_2 .
\]
Unless otherwise specified, we report the norm of the first loop output.

The gradient L2 norm is computed over all parameters of the first FFN block:
\[
\mathrm{GradNorm} =
\left(
\sum_{p \in \mathcal{P}_{\mathrm{FFN}_1}}
\|\nabla_p \mathcal{L}\|_2^2
\right)^{1/2}.
\]
We record this value before global gradient clipping.

\subsection{Architecture Details}
\label{app:arc}

\subsubsection*{Backbone Network Architecture}

All model variants share a common set of architectural choices.
We use Rotary Positional Embeddings (RoPE)~\citep{su2023roformerenhancedtransformerrotary} with base
$\theta = 10000$ and no learnable positional embeddings.
All normalization is performed by a parameter-free RMSNorm
(i.e.\ $\mathrm{norm}(\mathbf{x}) = \mathbf{x}/\|\mathbf{x}\|_{\mathrm{rms}}$
with no affine parameters).
Attention uses QK-Norm~\citep{henry2020querykeynormalizationtransformers}: after projecting queries and keys, each head vector is
normalized before computing attention weights, which improves training stability.
The FFN uses a $\mathrm{ReLU}^2$ activation~\citep{zhang2024relu2winsdiscoveringefficient} with an expansion ratio of 4.
Logits are soft-capped via $15\tanh(\ell / 15)$ before computing cross-entropy
loss, following~\cite{gemmateam2024gemma2improvingopen}.
The token embedding matrix and the language-model head are \emph{untied}
(separate parameters). No bias terms are used in any linear layer. Model dimensions are derived from a single \texttt{depth} hyperparameter:
\begin{equation}
  d_{\mathrm{model}} = 64 \times \texttt{depth}, \qquad
  H = \left\lceil d_{\mathrm{model}} / 128 \right\rceil, \qquad
  D = \texttt{depth},
\end{equation}
where $H$ is the number of query heads and $D$ is the number of unique
Transformer blocks. We use $K$ for the loop count in the experiments.
The vocabulary size is padded to the nearest multiple of 64 for computational
efficiency.

We use $D$ for the number of unique Transformer layers, $L$ for the number of training loops, and $K$ for the number of inference loops.
The total parameter count (standard MHA, padded vocab size $V$) is:
\begin{equation}
  N = 2Vd_{\mathrm{model}} + 12 D d_{\mathrm{model}}^2.
\end{equation}
Note that increasing the loop count $L$ adds compute but \emph{no} additional
parameters, since the $D$ blocks are reused across loop iterations.

\subsubsection*{Attention Variants}

We implement three orthogonal attention design axes, each selectable
independently: the KV projection scheme (\texttt{attn\_type}), the
head-sharing ratio (GQA), and the context window size
(\texttt{window\_pattern}).

\paragraph{Standard attention with Group-Query Attention (\texttt{attn\_type=``full''}).}
Each attention layer projects the input into $H_q$ query heads and $H_{kv}$
key/value heads:
\[
  \mathbf{W}_Q \in \mathbb{R}^{d \times H_q d_h}, \quad
  \mathbf{W}_K \in \mathbb{R}^{d \times H_{kv} d_h}, \quad
  \mathbf{W}_V \in \mathbb{R}^{d \times H_{kv} d_h}, \quad
  \mathbf{W}_O \in \mathbb{R}^{d \times d},
\]
where $d_h = d_{\mathrm{model}} / H_q$ is the per-head dimension.
RoPE is applied to each query and key head vector.
After RoPE, a parameter-free RMSNorm (QK-Norm) is applied independently to
each query and key head vector before computing attention weights.
This normalizes the pre-softmax dot products and improves training stability
without introducing learnable parameters.

Group-Query Attention requires $H_{kv} \mid H_q$: each of the $H_{kv}$
KV heads is broadcast to $H_q / H_{kv}$ query heads.
Setting $H_{kv} = H_q$ recovers standard Multi-Head Attention (MHA);
setting $H_{kv} = 1$ gives Multi-Query Attention (MQA). The per-block parameter count under GQA is $(10 + 2r)\,d_{\mathrm{model}}^2$ where $r = H_{kv}/H_q$,
compared to $12\,d_{\mathrm{model}}^2$ for MHA. We set $H_{kv} = H_q$ by default and apply GQA selectively as a KV-cache memory reduction technique.

\paragraph{Multi-head Latent Attention (\texttt{attn\_type=``mla''}).}
To further reduce KV-cache memory during inference, we implement a
low-rank KV compression scheme. Queries are projected in the standard way. For keys and values, a shared down-projection first compresses the input
to a low-dimensional latent:
\begin{equation}
  \mathbf{c}_{kv} = \mathrm{norm}\!\left(\mathbf{x}\,\mathbf{W}_{\mathrm{down}}\right),
  \qquad \mathbf{W}_{\mathrm{down}} \in \mathbb{R}^{d \times R},
\end{equation}
where $R$ is the KV latent rank (default $R = 128$) and $\mathrm{norm}$ is the
parameter-free RMSNorm.
The latent is then expanded back to full KV representations via separate
up-projections:
\[
  \mathbf{K} = \mathbf{c}_{kv}\,\mathbf{W}_{K}^{\uparrow}, \quad
  \mathbf{V} = \mathbf{c}_{kv}\,\mathbf{W}_{V}^{\uparrow}, \qquad
  \mathbf{W}_{K}^{\uparrow},\, \mathbf{W}_{V}^{\uparrow}
    \in \mathbb{R}^{R \times H_{kv} d_h}.
\]
RoPE and QK-Norm are applied to the expanded $\mathbf{Q}$ and $\mathbf{K}$
as in the standard case.
The RMSNorm on $\mathbf{c}_{kv}$ decouples the gradient flow between the
down-projection and the two up-projections, improving optimization stability.
The KV cache stores the expanded $\mathbf{K}$ and $\mathbf{V}$ tensors
(not the compressed latent), preserving the same inference interface as the
standard attention.
The per-block attention parameter count for MLA is
$d^2 + dR + 2R H_{kv} d_h + d^2$.
Because our implementation stores the expanded $\mathbf{K}$ and $\mathbf{V}$
tensors, it does not claim a reduced KV-cache footprint; the latent projection
instead changes the parameterization of the attention block.

\paragraph{Sliding Window Attention (\texttt{window\_pattern}).}
Independently of the KV projection scheme, each layer's attention span can be
restricted to a local window via the \texttt{window\_pattern} string.
The pattern is a sequence of characters (\texttt{L} or \texttt{S}) that is
tiled across the $D$ layers:
\begin{itemize}
  \item \texttt{L} (long): full causal context, $w = T$ (no restriction).
  \item \texttt{S} (short): local causal window of size $w = \lfloor T/4 \rfloor$.
\end{itemize}
The final layer is always forced to full context regardless of the pattern,
ensuring that the last-layer representations integrate global information.
Concretely, for query position $i$ and key position $j$, the attention mask
enforces both the causal constraint ($j \le i$) and, when applicable, the
window constraint ($i - j < w$).
When $w = T$ (full context), the fast \texttt{is\_causal=True} path of
\texttt{scaled\_dot\_product\_attention} is used; otherwise an explicit
Boolean mask is constructed and passed to the attention kernel.

\subsubsection*{Weight Initialization}

\begin{table}[h]
\centering
\caption{Weight initialization scheme.}
\begin{tabular}{lll}
\toprule
\textbf{Parameter} & \textbf{Distribution} & \textbf{Scale} \\
\midrule
Token embedding \texttt{wte} & Normal & $\sigma = 1.0$ \\
LM head \texttt{lm\_head}    & Normal & $\sigma = 0.001$ \\
$\mathbf{W}_Q,\mathbf{W}_K,\mathbf{W}_V$, FFN$_{\mathrm{fc}}$ & Uniform & $[-s,\, s],\; s = \sqrt{3/d_{\mathrm{model}}}$ \\
Output projections $\mathbf{W}_O$, FFN$_{\mathrm{proj}}$ & Constant & $0$ \\
\bottomrule
\end{tabular}
\end{table}

The Uniform bound $s = \sqrt{3/d_{\mathrm{model}}}$ achieves the same standard
deviation as $\mathcal{N}(0,\, 1/d_{\mathrm{model}})$.
Initializing output projections to zero ensures that the residual contribution
of each block is zero at the start of training.

\subsection{Experiments Details}
\label{app:exp}

We evaluate our models using language-modeling metrics and downstream accuracy.
Unless otherwise noted, all evaluations are run with bfloat16 mixed precision
under \texttt{torch.amp.autocast} and with a per-device batch size of 4 sequences.

\subsection*{Bits-per-Byte (BPB)}

During training we track validation BPB every 250 steps as the primary
per-training-step signal.
BPB is a tokenization-independent compression metric defined as
\begin{equation}
  \mathrm{BPB} = \frac{\sum_{t} \ell_t}{\log 2 \cdot \sum_{t} b_t},
\end{equation}
where $\ell_t$ is the per-token cross-entropy loss (in nats) and $b_t$ is the
number of UTF-8 bytes that token $t$ represents.
Special tokens (e.g.\ \texttt{<|im\_start|>}) contribute $b_t = 0$ and are
therefore excluded from both numerator and denominator, as are any positions
masked with \texttt{ignore\_index} $= -1$.
This normalization makes BPB comparable across models trained with different
vocabulary sizes.
BPB is reported on both the training corpus and the held-out validation shard
(the final Parquet shard).

\subsection*{CORE Benchmark}

CORE is an aggregate benchmark sourced from the DCLM evaluation suite.
It covers a diverse set of language understanding tasks grouped into three
evaluation types:

\begin{itemize}
  \item \textbf{Multiple-choice} (\texttt{multiple\_choice}):
        Each item has a shared context (query) and several candidate
        continuations.
        The model scores each candidate by computing the mean cross-entropy
        loss over the \emph{continuation-only} tokens (i.e.\ the tokens that
        differ across choices; the common prefix is not scored).
        The candidate with the lowest loss is selected as the prediction.

  \item \textbf{Schema} (\texttt{schema}):
        Each item has multiple possible contexts paired with a shared
        continuation.
        The model scores each context by computing the mean cross-entropy loss
        over the shared \emph{suffix} tokens.
        The context with the lowest loss is selected.

  \item \textbf{Language modeling} (\texttt{language\_modeling}):
        Each item has a single context and a target continuation.
        The model computes the mean cross-entropy loss over the continuation
        tokens only (identified via a prefix-length comparison).
        Correct prediction requires the loss to fall below a threshold implicitly
        defined by the accuracy computation.
\end{itemize}

The number of in-context examples is task-specific and fixed by the benchmark
configuration; the shot counts used for the downstream benchmarks are listed
in Table~\ref{tab:downstream_benchmarks}. Examples for each query are sampled without replacement using a
per-example random seed ($1234 + \text{idx}$), independently of the global
data shuffle.
Before evaluation, the full dataset for each task is shuffled with a fixed seed
($1337$) for reproducibility.
We use all available examples (i.e.\ \texttt{max\_per\_task} $= -1$) for
post-training evaluations; during training-time logging we subsample to at most
$500$ examples per task for efficiency.

The raw per-task accuracy $a$ is converted to a \emph{centered accuracy}:
\begin{equation}
  \tilde{a} = \frac{a - a_{\mathrm{rand}}}{1 - a_{\mathrm{rand}}},
\end{equation}
where $a_{\mathrm{rand}}$ is the random-baseline accuracy for that task
(provided by the benchmark metadata).
The overall \textbf{CORE metric} is the mean of all per-task centered accuracies. We use 21 benchmark task families; the logged aggregate contains 22 centered entries because HellaSwag is evaluated in both its standard and zero-shot configurations. Core values in Table~\ref{tab:main}, Table~\ref{tab:right}, and Figure~\ref{fig:tts} are displayed after multiplication by 100.

\subsection*{WikiText-2 Perplexity}

We evaluate standard language-model perplexity on the WikiText-2 validation
set:
\begin{equation}
  \mathrm{PPL} = \exp\!\left(\frac{1}{N}\sum_{t=1}^{N} \ell_t \right),
\end{equation}
where $\ell_t$ is the per-token NLL at position $t$ and $N$ is the total number
of non-masked tokens.
WikiText-2 PPL is always reported alongside CORE and BPB in post-training
evaluations.

\subsubsection*{Test-Time Compute Budget Evaluation}

A central aim of this work is to characterize how inference-time compute changes
performance when the loop count $K$ is adjusted without reloading or modifying
model weights. We run a \emph{budget evaluation} for the four Base-size FLT
checkpoints used in Figure~\ref{fig:tts}. Each checkpoint is evaluated at
$K \in \{1, 3, 6, 9, 12\}$ loop iterations in a single pass, logging BPB,
CORE, and the downstream-task accuracies at each budget point.
Results are recorded with $K$ as the x-axis, yielding a
compute--performance curve for each model.
This protocol compares each checkpoint at its training loop count with lower
and higher loop counts where available, and compares checkpoints at the same
inference loop count. Loop count is a compute proxy; latency and FLOPs are not
measured, and no matched-budget non-looped baseline is included. Figure~\ref{fig:tts}
contains all 20 settings from one checkpoint sweep per training loop count,
without repeated-seed uncertainty estimates. These observations do not establish
statistical significance or behavior beyond the evaluated range $K\leq12$.
Variable-loop training is not evaluated and remains future work. WikiText-2 PPL is
reported for post-training evaluations, but is not one of the metrics plotted
in Figure~\ref{fig:tts}.
\subsubsection*{Benchmark Statistics}

In addition to language-modeling metrics, we evaluate each trained model on a set of standard
downstream benchmarks. Table~\ref{tab:downstream_benchmarks} summarizes the benchmarks and
the number of in-context examples used for evaluation. We use 0-shot evaluation for LAMBADA OpenAI~\citep{paperno2016lambadadatasetwordprediction,radford2019language} and OpenBookQA~\citep{OpenBookQA2018}, and 10-shot evaluation for PIQA~\citep{Bisk2020}, HellaSwag~\citep{zellers2019hellaswag}, ARC-Easy~\citep{allenai:arc}, and ARC-Challenge~\citep{allenai:arc}. All downstream
results are reported as accuracy.

\vspace{0.1cm}
\begin{table}[h]
\centering
\small
\caption{
Downstream evaluation benchmarks used in this work.
The ``Shots'' column indicates the number of in-context examples used during evaluation.
All benchmarks are evaluated by selecting the answer with the highest model likelihood,
unless otherwise specified. We report accuracy for all downstream benchmarks.
}
\label{tab:downstream_benchmarks}
\vspace{0.1cm}
\resizebox{\textwidth}{!}{
\begin{tabular}{llllll}
\toprule
Benchmark & Shots & Task type \\
\midrule
LAMBADA OpenAI
& 0-shot
& Cloze-style language understanding \\

OpenBookQA
& 0-shot
& Multiple-choice open-book science question answering \\

PIQA
& 10-shot
& Physical commonsense reasoning \\

HellaSwag
& 10-shot
& Commonsense sentence completion \\

ARC-Easy
& 10-shot
& Multiple-choice science question answering \\

ARC-Challenge
& 10-shot
& Multiple-choice science question answering \\

\bottomrule
\end{tabular}
}
\end{table}

\section{Broader Impacts}
\label{app:imp}

This work is primarily methodological and studies how to improve the training stability of looped language models without increasing the number of learnable parameters. A potential positive impact is that more stable and parameter-efficient architectures may help reduce the cost of training and deploying language models, making test-time compute adjustment more accessible. By allowing performance to be traded against inference-time computation through the number of loop iterations, such models may also provide more flexible deployment options under different resource constraints.

At the same time, improvements in language-model training efficiency can have broader dual-use implications. More stable and compute-adaptive language models may lower the barrier to training capable generative systems, which could be misused for spam, misinformation, or other harmful content generation if deployed without safeguards. In addition, although looped models do not increase parameter count, using more loop iterations at inference increases computation and energy consumption. Like other language models trained on web-scale corpora, models based on this architecture may also inherit biases, factual errors, or harmful associations from the training data.

Our experiments are conducted on publicly available datasets and standard academic benchmarks, and this work does not release a high-risk pretrained model. Nevertheless, future applications of this architecture should be accompanied by appropriate safety evaluations, bias and robustness analyses, content-misuse mitigation, and transparent reporting of compute and energy costs.

\subsection{Existing Assets}
\label{app:assets}

We use publicly available benchmark datasets and standard open-source software libraries. The original sources of the datasets and any external code or models are cited in the main paper. We use these assets only for academic research and do not redistribute third-party datasets, models, or code.

\begin{table}[!h]
\centering
\small
\caption{Existing assets used in this work.}
\label{tab:existing_assets}
\begin{tabular}{l l l}
\toprule
Asset & Usage & License \\
\midrule
FineWeb-Edu     & Pretraining data              & ODC-BY-v1.0 \\
WikiText-2      & Perplexity evaluation         & CC-BY-SA-4.0 \\
PIQA            & Evaluation benchmark          & Not specified \\
HellaSwag       & Evaluation benchmark          & MIT \\
ARC             & Evaluation benchmark          & Not specified \\
OpenBookQA      & Evaluation benchmark          & Not specified \\
Qwen3 tokenizer & Tokenizer                     & Apache-2.0\\
nanochat        & Code reference                & MIT \\
\bottomrule
\end{tabular}
\end{table}

%% file: main.bbl
\begin{thebibliography}{44}
\providecommand{\natexlab}[1]{#1}
\providecommand{\url}[1]{\texttt{#1}}
\expandafter\ifx\csname urlstyle\endcsname\relax
  \providecommand{\doi}[1]{doi: #1}\else
  \providecommand{\doi}{doi: \begingroup \urlstyle{rm}\Url}\fi

\bibitem[Ainslie et~al.(2023)Ainslie, Lee-Thorp, de~Jong, Zemlyanskiy, Lebrón,
  and Sanghai]{ainslie2023gqatraininggeneralizedmultiquery}
Joshua Ainslie, James Lee-Thorp, Michiel de~Jong, Yury Zemlyanskiy, Federico
  Lebrón, and Sumit Sanghai.
\newblock Gqa: Training generalized multi-query transformer models from
  multi-head checkpoints, 2023.
\newblock URL \url{https://arxiv.org/abs/2305.13245}.

\bibitem[Anil et~al.(2022)Anil, Pokle, Liang, Treutlein, Wu, Bai, Kolter, and
  Grosse]{anil2022pat}
Cem Anil, Ashwini Pokle, Kaiqu Liang, Johannes Treutlein, Yuhuai Wu, Shaojie
  Bai, J~Zico Kolter, and Roger~B Grosse.
\newblock Path independent equilibrium models can better exploit test-time
  computation.
\newblock \emph{Advances in Neural Information Processing Systems},
  35:\penalty0 7796--7809, 2022.

\bibitem[Beltagy et~al.(2020)Beltagy, Peters, and
  Cohan]{beltagy2020longformerlongdocumentTransformer}
Iz~Beltagy, Matthew~E. Peters, and Arman Cohan.
\newblock Longformer: The long-document transformer, 2020.
\newblock URL \url{https://arxiv.org/abs/2004.05150}.

\bibitem[Bengio et~al.(1994)Bengio, Simard, and Frasconi]{bengio1994learning}
Yoshua Bengio, Patrice Simard, and Paolo Frasconi.
\newblock Learning long-term dependencies with gradient descent is difficult.
\newblock \emph{IEEE transactions on neural networks}, 5\penalty0 (2):\penalty0
  157--166, 1994.

\bibitem[Bisk et~al.(2020)Bisk, Zellers, Bras, Gao, and Choi]{Bisk2020}
Yonatan Bisk, Rowan Zellers, Ronan~Le Bras, Jianfeng Gao, and Yejin Choi.
\newblock Piqa: Reasoning about physical commonsense in natural language.
\newblock In \emph{Thirty-Fourth AAAI Conference on Artificial Intelligence},
  2020.

\bibitem[Cho et~al.(2014)Cho, Van~Merri{\"e}nboer, Gul{\c{c}}ehre, Bahdanau,
  Bougares, Schwenk, and Bengio]{cho2014learning}
Kyunghyun Cho, Bart Van~Merri{\"e}nboer, {\c{C}}a{\u{g}}lar Gul{\c{c}}ehre,
  Dzmitry Bahdanau, Fethi Bougares, Holger Schwenk, and Yoshua Bengio.
\newblock Learning phrase representations using rnn encoder--decoder for
  statistical machine translation.
\newblock In \emph{Proceedings of the 2014 conference on empirical methods in
  natural language processing (EMNLP)}, pp.\  1724--1734, 2014.

\bibitem[Clark et~al.(2018)Clark, Cowhey, Etzioni, Khot, Sabharwal, Schoenick,
  and Tafjord]{allenai:arc}
Peter Clark, Isaac Cowhey, Oren Etzioni, Tushar Khot, Ashish Sabharwal, Carissa
  Schoenick, and Oyvind Tafjord.
\newblock Think you have solved question answering? try arc, the ai2 reasoning
  challenge.
\newblock \emph{arXiv:1803.05457v1}, 2018.

\bibitem[DeepSeek-AI et~al.(2024)DeepSeek-AI, Liu, Feng, Wang, Wang, Liu, Zhao,
  Dengr, Ruan, Dai, Guo, Yang, Chen, Ji, Li, Lin, Luo, Hao, Chen, Li, Zhang,
  Xu, Yang, Zhang, Ding, Xin, Gao, Li, Qu, Cai, Liang, Guo, Ni, Li, Chen, Yuan,
  Qiu, Song, Dong, Gao, Guan, Wang, Zhang, Xu, Xia, Zhao, Zhang, Li, Wang,
  Zhang, Zhang, Tang, Li, Tian, Huang, Wang, Zhang, Zhu, Chen, Du, Chen, Jin,
  Ge, Pan, Xu, Chen, Li, Lu, Zhou, Chen, Wu, Ye, Ma, Wang, Zhou, Yu, Zhou,
  Zheng, Wang, Pei, Yuan, Sun, Xiao, Zeng, An, Liu, Liang, Gao, Zhang, Li, Jin,
  Wang, Bi, Liu, Wang, Shen, Chen, Chen, Nie, Sun, Wang, Liu, Xie, Yu, Song,
  Zhou, Yang, Lu, Su, Wu, Li, Wei, Zhu, Xu, Huang, Li, Zhao, Sun, Li, Wang,
  Zheng, Zhang, Xiong, Zhao, He, Tang, Piao, Dong, Tan, Liu, Wang, Guo, Zhu,
  Wang, Zou, Zha, Ma, Yan, You, Liu, Ren, Ren, Sha, Fu, Huang, Zhang, Xie, Hao,
  Shao, Wen, Xu, Zhang, Li, Wang, Gu, Li, and
  Xie]{deepseekai2024deepseekv2strongeconomicalefficient}
DeepSeek-AI, Aixin Liu, Bei Feng, Bin Wang, Bingxuan Wang, Bo~Liu, Chenggang
  Zhao, Chengqi Dengr, Chong Ruan, Damai Dai, Daya Guo, Dejian Yang, Deli Chen,
  Dongjie Ji, Erhang Li, Fangyun Lin, Fuli Luo, Guangbo Hao, Guanting Chen,
  Guowei Li, H.~Zhang, Hanwei Xu, Hao Yang, Haowei Zhang, Honghui Ding, Huajian
  Xin, Huazuo Gao, Hui Li, Hui Qu, J.~L. Cai, Jian Liang, Jianzhong Guo, Jiaqi
  Ni, Jiashi Li, Jin Chen, Jingyang Yuan, Junjie Qiu, Junxiao Song, Kai Dong,
  Kaige Gao, Kang Guan, Lean Wang, Lecong Zhang, Lei Xu, Leyi Xia, Liang Zhao,
  Liyue Zhang, Meng Li, Miaojun Wang, Mingchuan Zhang, Minghua Zhang, Minghui
  Tang, Mingming Li, Ning Tian, Panpan Huang, Peiyi Wang, Peng Zhang, Qihao
  Zhu, Qinyu Chen, Qiushi Du, R.~J. Chen, R.~L. Jin, Ruiqi Ge, Ruizhe Pan,
  Runxin Xu, Ruyi Chen, S.~S. Li, Shanghao Lu, Shangyan Zhou, Shanhuang Chen,
  Shaoqing Wu, Shengfeng Ye, Shirong Ma, Shiyu Wang, Shuang Zhou, Shuiping Yu,
  Shunfeng Zhou, Size Zheng, T.~Wang, Tian Pei, Tian Yuan, Tianyu Sun, W.~L.
  Xiao, Wangding Zeng, Wei An, Wen Liu, Wenfeng Liang, Wenjun Gao, Wentao
  Zhang, X.~Q. Li, Xiangyue Jin, Xianzu Wang, Xiao Bi, Xiaodong Liu, Xiaohan
  Wang, Xiaojin Shen, Xiaokang Chen, Xiaosha Chen, Xiaotao Nie, Xiaowen Sun,
  Xiaoxiang Wang, Xin Liu, Xin Xie, Xingkai Yu, Xinnan Song, Xinyi Zhou, Xinyu
  Yang, Xuan Lu, Xuecheng Su, Y.~Wu, Y.~K. Li, Y.~X. Wei, Y.~X. Zhu, Yanhong
  Xu, Yanping Huang, Yao Li, Yao Zhao, Yaofeng Sun, Yaohui Li, Yaohui Wang,
  Yi~Zheng, Yichao Zhang, Yiliang Xiong, Yilong Zhao, Ying He, Ying Tang, Yishi
  Piao, Yixin Dong, Yixuan Tan, Yiyuan Liu, Yongji Wang, Yongqiang Guo, Yuchen
  Zhu, Yuduan Wang, Yuheng Zou, Yukun Zha, Yunxian Ma, Yuting Yan, Yuxiang You,
  Yuxuan Liu, Z.~Z. Ren, Zehui Ren, Zhangli Sha, Zhe Fu, Zhen Huang, Zhen
  Zhang, Zhenda Xie, Zhewen Hao, Zhihong Shao, Zhiniu Wen, Zhipeng Xu, Zhongyu
  Zhang, Zhuoshu Li, Zihan Wang, Zihui Gu, Zilin Li, and Ziwei Xie.
\newblock Deepseek-v2: A strong, economical, and efficient mixture-of-experts
  language model, 2024.
\newblock URL \url{https://arxiv.org/abs/2405.04434}.

\bibitem[Dehghani et~al.(2019)Dehghani, Gouws, Vinyals, Uszkoreit, and
  Kaiser]{dehghaniuniversal}
Mostafa Dehghani, Stephan Gouws, Oriol Vinyals, Jakob Uszkoreit, and Lukasz
  Kaiser.
\newblock Universal transformers.
\newblock In \emph{International Conference on Learning Representations}, 2019.

\bibitem[Geiping et~al.(2025)Geiping, McLeish, Jain, Kirchenbauer, Singh,
  Bartoldson, Kailkhura, Bhatele, and Goldstein]{geiping2025scaling}
Jonas Geiping, Sean McLeish, Neel Jain, John Kirchenbauer, Siddharth Singh,
  Brian~R Bartoldson, Bhavya Kailkhura, Abhinav Bhatele, and Tom Goldstein.
\newblock Scaling up test-time compute with latent reasoning: A recurrent depth
  approach.
\newblock \emph{arXiv preprint arXiv:2502.05171}, 2025.

\bibitem[Giannou et~al.(2023)Giannou, Rajput, Sohn, Lee, Lee, and
  Papailiopoulos]{giannou2023looped}
Angeliki Giannou, Shashank Rajput, Jy-yong Sohn, Kangwook Lee, Jason~D Lee, and
  Dimitris Papailiopoulos.
\newblock Looped transformers as programmable computers.
\newblock In \emph{International Conference on Machine Learning}, pp.\
  11398--11442. PMLR, 2023.

\bibitem[Guo et~al.(2025)Guo, Yang, Zhang, Song, Wang, Zhu, Xu, Zhang, Ma, Bi,
  et~al.]{guo2025deepseek}
Daya Guo, Dejian Yang, Haowei Zhang, Junxiao Song, Peiyi Wang, Qihao Zhu,
  Runxin Xu, Ruoyu Zhang, Shirong Ma, Xiao Bi, et~al.
\newblock Deepseek-r1: Incentivizing reasoning capability in llms via
  reinforcement learning.
\newblock \emph{arXiv preprint arXiv:2501.12948}, 2025.

\bibitem[He et~al.(2016)He, Zhang, Ren, and Sun]{he2016deep}
Kaiming He, Xiangyu Zhang, Shaoqing Ren, and Jian Sun.
\newblock Deep residual learning for image recognition.
\newblock In \emph{Proceedings of the IEEE conference on computer vision and
  pattern recognition}, pp.\  770--778, 2016.

\bibitem[Henry et~al.(2020)Henry, Dachapally, Pawar, and
  Chen]{henry2020querykeynormalizationtransformers}
Alex Henry, Prudhvi~Raj Dachapally, Shubham Pawar, and Yuxuan Chen.
\newblock Query-key normalization for transformers, 2020.
\newblock URL \url{https://arxiv.org/abs/2010.04245}.

\bibitem[Hochreiter \& Schmidhuber(1997)Hochreiter and
  Schmidhuber]{hochreiter1997long}
Sepp Hochreiter and J{\"u}rgen Schmidhuber.
\newblock Long short-term memory.
\newblock \emph{Neural computation}, 9\penalty0 (8):\penalty0 1735--1780, 1997.

\bibitem[Hoffmann et~al.(2022)Hoffmann, Borgeaud, Mensch, Buchatskaya, Cai,
  Rutherford, Casas, Hendricks, Welbl, Clark, et~al.]{hoffmann2022training}
Jordan Hoffmann, Sebastian Borgeaud, Arthur Mensch, Elena Buchatskaya, Trevor
  Cai, Eliza Rutherford, DDL Casas, Lisa~Anne Hendricks, Johannes Welbl, Aidan
  Clark, et~al.
\newblock Training compute-optimal large language models.
\newblock \emph{arXiv preprint arXiv:2203.15556}, 10, 2022.

\bibitem[Jordan et~al.(2024{\natexlab{a}})Jordan, Bernstein, Rappazzo,
  @fernbear.bsky.social, Vlado, Jiacheng, Cesista, Koszarsky, and
  @Grad62304977]{modded_nanogpt_2024}
Keller Jordan, Jeremy Bernstein, Brendan Rappazzo, @fernbear.bsky.social, Boza
  Vlado, You Jiacheng, Franz Cesista, Braden Koszarsky, and @Grad62304977.
\newblock modded-nanogpt: Speedrunning the nanogpt baseline,
  2024{\natexlab{a}}.
\newblock URL \url{https://github.com/KellerJordan/modded-nanogpt}.

\bibitem[Jordan et~al.(2024{\natexlab{b}})Jordan, Jin, Boza, You, Cesista,
  Newhouse, and Bernstein]{jordan2024muon}
Keller Jordan, Yuchen Jin, Vlado Boza, Jiacheng You, Franz Cesista, Laker
  Newhouse, and Jeremy Bernstein.
\newblock Muon: An optimizer for hidden layers in neural networks,
  2024{\natexlab{b}}.
\newblock URL \url{https://kellerjordan.github.io/posts/muon/}.

\bibitem[Karpathy(2025)]{nanochat}
Andrej Karpathy.
\newblock nanochat: The best chatgpt that \$100 can buy, 2025.
\newblock URL \url{https://github.com/karpathy/nanochat}.

\bibitem[Koishekenov et~al.(2025)Koishekenov, Lipani, and
  Cancedda]{koishekenov2025encode}
Yeskendir Koishekenov, Aldo Lipani, and Nicola Cancedda.
\newblock Encode, think, decode: Scaling test-time reasoning with recursive
  latent thoughts.
\newblock \emph{arXiv preprint arXiv:2510.07358}, 2025.

\bibitem[Li et~al.(2024)Li, Fang, Smyrnis, Ivgi, Jordan, Gadre, Bansal, Guha,
  Keh, Arora, et~al.]{li2024datacomp}
Jeffrey Li, Alex Fang, Georgios Smyrnis, Maor Ivgi, Matt Jordan, Samir Gadre,
  Hritik Bansal, Etash Guha, Sedrick Keh, Kushal Arora, et~al.
\newblock Datacomp-lm: In search of the next generation of training sets for
  language models.
\newblock \emph{Advances in Neural Information Processing Systems},
  37:\penalty0 14200--14282, 2024.

\bibitem[Loshchilov \& Hutter(2019)Loshchilov and
  Hutter]{loshchilov2019decoupledweightdecayregularization}
Ilya Loshchilov and Frank Hutter.
\newblock Decoupled weight decay regularization, 2019.
\newblock URL \url{https://arxiv.org/abs/1711.05101}.

\bibitem[Lozhkov et~al.(2024)Lozhkov, Ben~Allal, von Werra, and
  Wolf]{lozhkov2024fineweb-edu}
Anton Lozhkov, Loubna Ben~Allal, Leandro von Werra, and Thomas Wolf.
\newblock Fineweb-edu: the finest collection of educational content, 2024.
\newblock URL \url{https://huggingface.co/datasets/HuggingFaceFW/fineweb-edu}.

\bibitem[Merity et~al.(2016)Merity, Xiong, Bradbury, and
  Socher]{merity2016pointer}
Stephen Merity, Caiming Xiong, James Bradbury, and Richard Socher.
\newblock Pointer sentinel mixture models, 2016.

\bibitem[Mihaylov et~al.(2018)Mihaylov, Clark, Khot, and
  Sabharwal]{OpenBookQA2018}
Todor Mihaylov, Peter Clark, Tushar Khot, and Ashish Sabharwal.
\newblock Can a suit of armor conduct electricity? a new dataset for open book
  question answering.
\newblock In \emph{EMNLP}, 2018.

\bibitem[Paperno et~al.(2016)Paperno, Kruszewski, Lazaridou, Pham, Bernardi,
  Pezzelle, Baroni, Boleda, and
  Fernández]{paperno2016lambadadatasetwordprediction}
Denis Paperno, Germán Kruszewski, Angeliki Lazaridou, Quan~Ngoc Pham,
  Raffaella Bernardi, Sandro Pezzelle, Marco Baroni, Gemma Boleda, and Raquel
  Fernández.
\newblock The lambada dataset: Word prediction requiring a broad discourse
  context, 2016.
\newblock URL \url{https://arxiv.org/abs/1606.06031}.

\bibitem[Pascanu et~al.(2013)Pascanu, Mikolov, and
  Bengio]{pascanu2013difficulty}
Razvan Pascanu, Tomas Mikolov, and Yoshua Bengio.
\newblock On the difficulty of training recurrent neural networks.
\newblock In \emph{International conference on machine learning}, pp.\
  1310--1318. Pmlr, 2013.

\bibitem[Radford et~al.(2019)Radford, Wu, Child, Luan, Amodei, and
  Sutskever]{radford2019language}
Alec Radford, Jeff Wu, Rewon Child, David Luan, Dario Amodei, and Ilya
  Sutskever.
\newblock Language models are unsupervised multitask learners.
\newblock 2019.

\bibitem[Rajbhandari et~al.(2020)Rajbhandari, Rasley, Ruwase, and
  He]{rajbhandari2020zeromemoryoptimizationstraining}
Samyam Rajbhandari, Jeff Rasley, Olatunji Ruwase, and Yuxiong He.
\newblock Zero: Memory optimizations toward training trillion parameter models,
  2020.
\newblock URL \url{https://arxiv.org/abs/1910.02054}.

\bibitem[Saunshi et~al.(2025)Saunshi, Dikkala, Li, Kumar, and
  Reddi]{saunshi2025reasoning}
Nikunj Saunshi, Nishanth Dikkala, Zhiyuan Li, Sanjiv Kumar, and Sashank~J
  Reddi.
\newblock Reasoning with latent thoughts: On the power of looped transformers.
\newblock \emph{arXiv preprint arXiv:2502.17416}, 2025.

\bibitem[Su et~al.(2023)Su, Lu, Pan, Murtadha, Wen, and
  Liu]{su2023roformerenhancedtransformerrotary}
Jianlin Su, Yu~Lu, Shengfeng Pan, Ahmed Murtadha, Bo~Wen, and Yunfeng Liu.
\newblock Roformer: Enhanced transformer with rotary position embedding, 2023.
\newblock URL \url{https://arxiv.org/abs/2104.09864}.

\bibitem[Team et~al.(2024)Team, Riviere, Pathak, Sessa, Hardin, Bhupatiraju,
  Hussenot, Mesnard, Shahriari, Ramé, Ferret, Liu, Tafti, Friesen, Casbon,
  Ramos, Kumar, Lan, Jerome, Tsitsulin, Vieillard, Stanczyk, Girgin, Momchev,
  Hoffman, Thakoor, Grill, Neyshabur, Bachem, Walton, Severyn, Parrish, Ahmad,
  Hutchison, Abdagic, Carl, Shen, Brock, Coenen, Laforge, Paterson, Bastian,
  Piot, Wu, Royal, Chen, Kumar, Perry, Welty, Choquette-Choo, Sinopalnikov,
  Weinberger, Vijaykumar, Rogozińska, Herbison, Bandy, Wang, Noland, Moreira,
  Senter, Eltyshev, Visin, Rasskin, Wei, Cameron, Martins, Hashemi,
  Klimczak-Plucińska, Batra, Dhand, Nardini, Mein, Zhou, Svensson, Stanway,
  Chan, Zhou, Carrasqueira, Iljazi, Becker, Fernandez, van Amersfoort, Gordon,
  Lipschultz, Newlan, yeong Ji, Mohamed, Badola, Black, Millican, McDonell,
  Nguyen, Sodhia, Greene, Sjoesund, Usui, Sifre, Heuermann, Lago, McNealus,
  Soares, Kilpatrick, Dixon, Martins, Reid, Singh, Iverson, Görner, Velloso,
  Wirth, Davidow, Miller, Rahtz, Watson, Risdal, Kazemi, Moynihan, Zhang,
  Kahng, Park, Rahman, Khatwani, Dao, Bardoliwalla, Devanathan, Dumai, Chauhan,
  Wahltinez, Botarda, Barnes, Barham, Michel, Jin, Georgiev, Culliton, Kuppala,
  Comanescu, Merhej, Jana, Rokni, Agarwal, Mullins, Saadat, Carthy, Cogan,
  Perrin, Arnold, Krause, Dai, Garg, Sheth, Ronstrom, Chan, Jordan, Yu, Eccles,
  Hennigan, Kocisky, Doshi, Jain, Yadav, Meshram, Dharmadhikari, Barkley, Wei,
  Ye, Han, Kwon, Xu, Shen, Gong, Wei, Cotruta, Kirk, Rao, Giang, Peran,
  Warkentin, Collins, Barral, Ghahramani, Hadsell, Sculley, Banks, Dragan,
  Petrov, Vinyals, Dean, Hassabis, Kavukcuoglu, Farabet, Buchatskaya, Borgeaud,
  Fiedel, Joulin, Kenealy, Dadashi, and
  Andreev]{gemmateam2024gemma2improvingopen}
Gemma Team, Morgane Riviere, Shreya Pathak, Pier~Giuseppe Sessa, Cassidy
  Hardin, Surya Bhupatiraju, Léonard Hussenot, Thomas Mesnard, Bobak
  Shahriari, Alexandre Ramé, Johan Ferret, Peter Liu, Pouya Tafti, Abe
  Friesen, Michelle Casbon, Sabela Ramos, Ravin Kumar, Charline~Le Lan, Sammy
  Jerome, Anton Tsitsulin, Nino Vieillard, Piotr Stanczyk, Sertan Girgin,
  Nikola Momchev, Matt Hoffman, Shantanu Thakoor, Jean-Bastien Grill, Behnam
  Neyshabur, Olivier Bachem, Alanna Walton, Aliaksei Severyn, Alicia Parrish,
  Aliya Ahmad, Allen Hutchison, Alvin Abdagic, Amanda Carl, Amy Shen, Andy
  Brock, Andy Coenen, Anthony Laforge, Antonia Paterson, Ben Bastian, Bilal
  Piot, Bo~Wu, Brandon Royal, Charlie Chen, Chintu Kumar, Chris Perry, Chris
  Welty, Christopher~A. Choquette-Choo, Danila Sinopalnikov, David Weinberger,
  Dimple Vijaykumar, Dominika Rogozińska, Dustin Herbison, Elisa Bandy, Emma
  Wang, Eric Noland, Erica Moreira, Evan Senter, Evgenii Eltyshev, Francesco
  Visin, Gabriel Rasskin, Gary Wei, Glenn Cameron, Gus Martins, Hadi Hashemi,
  Hanna Klimczak-Plucińska, Harleen Batra, Harsh Dhand, Ivan Nardini, Jacinda
  Mein, Jack Zhou, James Svensson, Jeff Stanway, Jetha Chan, Jin~Peng Zhou,
  Joana Carrasqueira, Joana Iljazi, Jocelyn Becker, Joe Fernandez, Joost van
  Amersfoort, Josh Gordon, Josh Lipschultz, Josh Newlan, Ju~yeong Ji, Kareem
  Mohamed, Kartikeya Badola, Kat Black, Katie Millican, Keelin McDonell, Kelvin
  Nguyen, Kiranbir Sodhia, Kish Greene, Lars~Lowe Sjoesund, Lauren Usui,
  Laurent Sifre, Lena Heuermann, Leticia Lago, Lilly McNealus, Livio~Baldini
  Soares, Logan Kilpatrick, Lucas Dixon, Luciano Martins, Machel Reid,
  Manvinder Singh, Mark Iverson, Martin Görner, Mat Velloso, Mateo Wirth, Matt
  Davidow, Matt Miller, Matthew Rahtz, Matthew Watson, Meg Risdal, Mehran
  Kazemi, Michael Moynihan, Ming Zhang, Minsuk Kahng, Minwoo Park, Mofi Rahman,
  Mohit Khatwani, Natalie Dao, Nenshad Bardoliwalla, Nesh Devanathan, Neta
  Dumai, Nilay Chauhan, Oscar Wahltinez, Pankil Botarda, Parker Barnes, Paul
  Barham, Paul Michel, Pengchong Jin, Petko Georgiev, Phil Culliton, Pradeep
  Kuppala, Ramona Comanescu, Ramona Merhej, Reena Jana, Reza~Ardeshir Rokni,
  Rishabh Agarwal, Ryan Mullins, Samaneh Saadat, Sara~Mc Carthy, Sarah Cogan,
  Sarah Perrin, Sébastien M.~R. Arnold, Sebastian Krause, Shengyang Dai,
  Shruti Garg, Shruti Sheth, Sue Ronstrom, Susan Chan, Timothy Jordan, Ting Yu,
  Tom Eccles, Tom Hennigan, Tomas Kocisky, Tulsee Doshi, Vihan Jain, Vikas
  Yadav, Vilobh Meshram, Vishal Dharmadhikari, Warren Barkley, Wei Wei, Wenming
  Ye, Woohyun Han, Woosuk Kwon, Xiang Xu, Zhe Shen, Zhitao Gong, Zichuan Wei,
  Victor Cotruta, Phoebe Kirk, Anand Rao, Minh Giang, Ludovic Peran, Tris
  Warkentin, Eli Collins, Joelle Barral, Zoubin Ghahramani, Raia Hadsell,
  D.~Sculley, Jeanine Banks, Anca Dragan, Slav Petrov, Oriol Vinyals, Jeff
  Dean, Demis Hassabis, Koray Kavukcuoglu, Clement Farabet, Elena Buchatskaya,
  Sebastian Borgeaud, Noah Fiedel, Armand Joulin, Kathleen Kenealy, Robert
  Dadashi, and Alek Andreev.
\newblock Gemma 2: Improving open language models at a practical size, 2024.
\newblock URL \url{https://arxiv.org/abs/2408.00118}.

\bibitem[team et~al.(2024)team, Jaech, Kalai, Lerer, Richardson, El-Kishky,
  Low, Helyar, Madry, Beutel, Carney, Iftimie, Karpenko, Passos, Neitz,
  Prokofiev, Wei, Tam, Bennett, Kumar, Saraiva, Vallone, Duberstein, Kondrich,
  Mishchenko, Applebaum, Jiang, Nair, Zoph, Ghorbani, Rossen, Sokolowsky,
  Barak, McGrew, Minaiev, Hao, Baker, Houghton, McKinzie, Eastman, Lugaresi,
  Bassin, Hudson, Li, de~Bourcy, Voss, Shen, Zhang, Koch, Orsinger, Hesse,
  Fischer, Chan, Roberts, Kappler, Levy, Selsam, Dohan, Farhi, Mely, Robinson,
  Tsipras, Li, Oprica, Freeman, Zhang, Wong, Proehl, Cheung, Mitchell, Wallace,
  Ritter, Mays, Wang, Such, Raso, Leoni, Tsimpourlas, Song, von Lohmann, Sulit,
  Salmon, Parascandolo, Chabot, Zhao, Brockman, Leclerc, Salman, Bao, Sheng,
  Andrin, Bagherinezhad, Ren, Lightman, Chung, Kivlichan, O'Connell, Osband,
  Gilaberte, Akkaya, Kostrikov, Sutskever, Kofman, Pachocki, Lennon, Wei, Harb,
  Twore, Feng, Yu, Weng, Tang, Yu, Candela, Palermo, Parish, Heidecke, Hallman,
  Rizzo, Gordon, Uesato, Ward, Huizinga, Wang, Chen, Xiao, Singhal, Nguyen,
  Cobbe, Shi, Wood, Rimbach, Gu-Lemberg, Liu, Lu, Stone, Yu, Ahmad, Yang, Liu,
  Maksin, Ho, Fedus, Weng, Li, McCallum, Held, Kuhn, Kondraciuk, Kaiser, Metz,
  Boyd, Trebacz, Joglekar, Chen, Tintor, Meyer, Jones, Kaufer, Schwarzer, Shah,
  Yatbaz, Guan, Xu, Yan, Glaese, Chen, Lampe, Malek, Wang, Fradin, McClay,
  Pavlov, Wang, Wang, Murati, Bavarian, Rohaninejad, McAleese, Chowdhury,
  Chowdhury, Ryder, Tezak, Brown, Nachum, Boiko, Murk, Watkins, Chao,
  Ashbourne, Izmailov, Zhokhov, Dias, Arora, Lin, Lopes, Gaon, Miyara, Leike,
  Hwang, Garg, Brown, James, Shu, Cheu, Greene, Jain, Altman, Toizer, Toyer,
  Miserendino, Agarwal, Hernandez, Baker, McKinney, Yan, Zhao, Hu, Santurkar,
  Chaudhuri, Zhang, Fu, Papay, Lin, Balaji, Sanjeev, Sidor, Broda, Clark, Wang,
  Gordon, Sanders, Patwardhan, Sottiaux, Degry, Dimson, Zheng, Garipov, Stasi,
  Bansal, Creech, Peterson, Eloundou, Qi, Kosaraju, Monaco, Pong, Fomenko,
  Zheng, Zhou, McCabe, Zaremba, Dubois, Lu, Chen, Cha, Bai, He, Zhang, Wang,
  Shao, and Li]{openai2024openaio1card}
OpenAI team, Aaron Jaech, Adam Kalai, Adam Lerer, Adam Richardson, Ahmed
  El-Kishky, Aiden Low, Alec Helyar, Aleksander Madry, Alex Beutel, Alex
  Carney, Alex Iftimie, Alex Karpenko, Alex~Tachard Passos, Alexander Neitz,
  Alexander Prokofiev, Alexander Wei, Allison Tam, Ally Bennett, Ananya Kumar,
  Andre Saraiva, Andrea Vallone, Andrew Duberstein, Andrew Kondrich, Andrey
  Mishchenko, Andy Applebaum, Angela Jiang, Ashvin Nair, Barret Zoph, Behrooz
  Ghorbani, Ben Rossen, Benjamin Sokolowsky, Boaz Barak, Bob McGrew, Borys
  Minaiev, Botao Hao, Bowen Baker, Brandon Houghton, Brandon McKinzie, Brydon
  Eastman, Camillo Lugaresi, Cary Bassin, Cary Hudson, Chak~Ming Li, Charles
  de~Bourcy, Chelsea Voss, Chen Shen, Chong Zhang, Chris Koch, Chris Orsinger,
  Christopher Hesse, Claudia Fischer, Clive Chan, Dan Roberts, Daniel Kappler,
  Daniel Levy, Daniel Selsam, David Dohan, David Farhi, David Mely, David
  Robinson, Dimitris Tsipras, Doug Li, Dragos Oprica, Eben Freeman, Eddie
  Zhang, Edmund Wong, Elizabeth Proehl, Enoch Cheung, Eric Mitchell, Eric
  Wallace, Erik Ritter, Evan Mays, Fan Wang, Felipe~Petroski Such, Filippo
  Raso, Florencia Leoni, Foivos Tsimpourlas, Francis Song, Fred von Lohmann,
  Freddie Sulit, Geoff Salmon, Giambattista Parascandolo, Gildas Chabot, Grace
  Zhao, Greg Brockman, Guillaume Leclerc, Hadi Salman, Haiming Bao, Hao Sheng,
  Hart Andrin, Hessam Bagherinezhad, Hongyu Ren, Hunter Lightman, Hyung~Won
  Chung, Ian Kivlichan, Ian O'Connell, Ian Osband, Ignasi~Clavera Gilaberte,
  Ilge Akkaya, Ilya Kostrikov, Ilya Sutskever, Irina Kofman, Jakub Pachocki,
  James Lennon, Jason Wei, Jean Harb, Jerry Twore, Jiacheng Feng, Jiahui Yu,
  Jiayi Weng, Jie Tang, Jieqi Yu, Joaquin~Quiñonero Candela, Joe Palermo, Joel
  Parish, Johannes Heidecke, John Hallman, John Rizzo, Jonathan Gordon,
  Jonathan Uesato, Jonathan Ward, Joost Huizinga, Julie Wang, Kai Chen, Kai
  Xiao, Karan Singhal, Karina Nguyen, Karl Cobbe, Katy Shi, Kayla Wood, Kendra
  Rimbach, Keren Gu-Lemberg, Kevin Liu, Kevin Lu, Kevin Stone, Kevin Yu, Lama
  Ahmad, Lauren Yang, Leo Liu, Leon Maksin, Leyton Ho, Liam Fedus, Lilian Weng,
  Linden Li, Lindsay McCallum, Lindsey Held, Lorenz Kuhn, Lukas Kondraciuk,
  Lukasz Kaiser, Luke Metz, Madelaine Boyd, Maja Trebacz, Manas Joglekar, Mark
  Chen, Marko Tintor, Mason Meyer, Matt Jones, Matt Kaufer, Max Schwarzer,
  Meghan Shah, Mehmet Yatbaz, Melody~Y. Guan, Mengyuan Xu, Mengyuan Yan, Mia
  Glaese, Mianna Chen, Michael Lampe, Michael Malek, Michele Wang, Michelle
  Fradin, Mike McClay, Mikhail Pavlov, Miles Wang, Mingxuan Wang, Mira Murati,
  Mo~Bavarian, Mostafa Rohaninejad, Nat McAleese, Neil Chowdhury, Neil
  Chowdhury, Nick Ryder, Nikolas Tezak, Noam Brown, Ofir Nachum, Oleg Boiko,
  Oleg Murk, Olivia Watkins, Patrick Chao, Paul Ashbourne, Pavel Izmailov,
  Peter Zhokhov, Rachel Dias, Rahul Arora, Randall Lin, Rapha~Gontijo Lopes,
  Raz Gaon, Reah Miyara, Reimar Leike, Renny Hwang, Rhythm Garg, Robin Brown,
  Roshan James, Rui Shu, Ryan Cheu, Ryan Greene, Saachi Jain, Sam Altman, Sam
  Toizer, Sam Toyer, Samuel Miserendino, Sandhini Agarwal, Santiago Hernandez,
  Sasha Baker, Scott McKinney, Scottie Yan, Shengjia Zhao, Shengli Hu, Shibani
  Santurkar, Shraman~Ray Chaudhuri, Shuyuan Zhang, Siyuan Fu, Spencer Papay,
  Steph Lin, Suchir Balaji, Suvansh Sanjeev, Szymon Sidor, Tal Broda, Aidan
  Clark, Tao Wang, Taylor Gordon, Ted Sanders, Tejal Patwardhan, Thibault
  Sottiaux, Thomas Degry, Thomas Dimson, Tianhao Zheng, Timur Garipov, Tom
  Stasi, Trapit Bansal, Trevor Creech, Troy Peterson, Tyna Eloundou, Valerie
  Qi, Vineet Kosaraju, Vinnie Monaco, Vitchyr Pong, Vlad Fomenko, Weiyi Zheng,
  Wenda Zhou, Wes McCabe, Wojciech Zaremba, Yann Dubois, Yinghai Lu, Yining
  Chen, Young Cha, Yu~Bai, Yuchen He, Yuchen Zhang, Yunyun Wang, Zheng Shao,
  and Zhuohan Li.
\newblock Openai o1 system card, 2024.
\newblock URL \url{https://arxiv.org/abs/2412.16720}.

\bibitem[Vaswani et~al.(2023)Vaswani, Shazeer, Parmar, Uszkoreit, Jones, Gomez,
  Kaiser, and Polosukhin]{vaswani2023attentionneed}
Ashish Vaswani, Noam Shazeer, Niki Parmar, Jakob Uszkoreit, Llion Jones,
  Aidan~N. Gomez, Lukasz Kaiser, and Illia Polosukhin.
\newblock Attention is all you need, 2023.
\newblock URL \url{https://arxiv.org/abs/1706.03762}.

\bibitem[Veit et~al.(2016)Veit, Wilber, and Belongie]{veit2016residual}
Andreas Veit, Michael~J Wilber, and Serge Belongie.
\newblock Residual networks behave like ensembles of relatively shallow
  networks.
\newblock \emph{Advances in neural information processing systems}, 29, 2016.

\bibitem[Villalobos et~al.(2024)Villalobos, Ho, Sevilla, Besiroglu, Heim, and
  Hobbhahn]{villalobos2024position}
Pablo Villalobos, Anson Ho, Jaime Sevilla, Tamay Besiroglu, Lennart Heim, and
  Marius Hobbhahn.
\newblock Position: Will we run out of data? limits of llm scaling based on
  human-generated data.
\newblock In \emph{Forty-first International Conference on Machine Learning},
  2024.

\bibitem[Wei et~al.(2023)Wei, Wang, Schuurmans, Bosma, Ichter, Xia, Chi, Le,
  and Zhou]{wei2023chainofthoughtpromptingelicitsreasoning}
Jason Wei, Xuezhi Wang, Dale Schuurmans, Maarten Bosma, Brian Ichter, Fei Xia,
  Ed~Chi, Quoc Le, and Denny Zhou.
\newblock Chain-of-thought prompting elicits reasoning in large language
  models, 2023.
\newblock URL \url{https://arxiv.org/abs/2201.11903}.

\bibitem[Werbos(2002)]{werbos2002backpropagation}
Paul~J Werbos.
\newblock Backpropagation through time: what it does and how to do it.
\newblock \emph{Proceedings of the IEEE}, 78\penalty0 (10):\penalty0
  1550--1560, 2002.

\bibitem[Yang et~al.(2025)Yang, Li, Yang, Zhang, Hui, Zheng, Yu, Gao, Huang,
  Lv, Zheng, Liu, Zhou, Huang, Hu, Ge, Wei, Lin, Tang, Yang, Tu, Zhang, Yang,
  Yang, Zhou, Zhou, Lin, Dang, Bao, Yang, Yu, Deng, Li, Xue, Li, Zhang, Wang,
  Zhu, Men, Gao, Liu, Luo, Li, Tang, Yin, Ren, Wang, Zhang, Ren, Fan, Su,
  Zhang, Zhang, Wan, Liu, Wang, Cui, Zhang, Zhou, and
  Qiu]{yang2025qwen3technicalreport}
An~Yang, Anfeng Li, Baosong Yang, Beichen Zhang, Binyuan Hui, Bo~Zheng, Bowen
  Yu, Chang Gao, Chengen Huang, Chenxu Lv, Chujie Zheng, Dayiheng Liu, Fan
  Zhou, Fei Huang, Feng Hu, Hao Ge, Haoran Wei, Huan Lin, Jialong Tang, Jian
  Yang, Jianhong Tu, Jianwei Zhang, Jianxin Yang, Jiaxi Yang, Jing Zhou,
  Jingren Zhou, Junyang Lin, Kai Dang, Keqin Bao, Kexin Yang, Le~Yu, Lianghao
  Deng, Mei Li, Mingfeng Xue, Mingze Li, Pei Zhang, Peng Wang, Qin Zhu, Rui
  Men, Ruize Gao, Shixuan Liu, Shuang Luo, Tianhao Li, Tianyi Tang, Wenbiao
  Yin, Xingzhang Ren, Xinyu Wang, Xinyu Zhang, Xuancheng Ren, Yang Fan, Yang
  Su, Yichang Zhang, Yinger Zhang, Yu~Wan, Yuqiong Liu, Zekun Wang, Zeyu Cui,
  Zhenru Zhang, Zhipeng Zhou, and Zihan Qiu.
\newblock Qwen3 technical report, 2025.
\newblock URL \url{https://arxiv.org/abs/2505.09388}.

\bibitem[Yang et~al.(2022)Yang, Hu, Babuschkin, Sidor, Liu, Farhi, Ryder,
  Pachocki, Chen, and Gao]{yang2022tensorprogramsvtuning}
Greg Yang, Edward~J. Hu, Igor Babuschkin, Szymon Sidor, Xiaodong Liu, David
  Farhi, Nick Ryder, Jakub Pachocki, Weizhu Chen, and Jianfeng Gao.
\newblock Tensor programs v: Tuning large neural networks via zero-shot
  hyperparameter transfer, 2022.
\newblock URL \url{https://arxiv.org/abs/2203.03466}.

\bibitem[Zellers et~al.(2019)Zellers, Holtzman, Bisk, Farhadi, and
  Choi]{zellers2019hellaswag}
Rowan Zellers, Ari Holtzman, Yonatan Bisk, Ali Farhadi, and Yejin Choi.
\newblock Hellaswag: Can a machine really finish your sentence?
\newblock In \emph{Proceedings of the 57th Annual Meeting of the Association
  for Computational Linguistics}, 2019.

\bibitem[Zhang \& Sennrich(2019)Zhang and
  Sennrich]{zhang2019rootmeansquarelayer}
Biao Zhang and Rico Sennrich.
\newblock Root mean square layer normalization, 2019.
\newblock URL \url{https://arxiv.org/abs/1910.07467}.

\bibitem[Zhang et~al.(2024)Zhang, Song, Yu, Han, Lin, Xiao, Song, Liu, Mi, and
  Sun]{zhang2024relu2winsdiscoveringefficient}
Zhengyan Zhang, Yixin Song, Guanghui Yu, Xu~Han, Yankai Lin, Chaojun Xiao,
  Chenyang Song, Zhiyuan Liu, Zeyu Mi, and Maosong Sun.
\newblock Relu$^2$ wins: Discovering efficient activation functions for sparse
  llms, 2024.
\newblock URL \url{https://arxiv.org/abs/2402.03804}.

\bibitem[Zhu et~al.(2025)Zhu, Wang, Hua, Zhang, Li, Que, Wei, Wen, Yin, Xing,
  et~al.]{zhu2025scaling}
Rui-Jie Zhu, Zixuan Wang, Kai Hua, Tianyu Zhang, Ziniu Li, Haoran Que, Boyi
  Wei, Zixin Wen, Fan Yin, He~Xing, et~al.
\newblock Scaling latent reasoning via looped language models.
\newblock \emph{arXiv preprint arXiv:2510.25741}, 2025.

\end{thebibliography}
